\documentclass{article} 
\usepackage{iclr2025_conference,times}

\usepackage{amsmath,amsfonts,bm}

\def\eqref#1{equation~\ref{#1}}

\def\1{\bm{1}}

\DeclareMathAlphabet{\mathsfit}{\encodingdefault}{\sfdefault}{m}{sl}
\SetMathAlphabet{\mathsfit}{bold}{\encodingdefault}{\sfdefault}{bx}{n}

\usepackage{hyperref}
\hypersetup{
    colorlinks=false,    
    pdfborder={0 0 0}    
}
\usepackage{url}
\usepackage[OT1]{fontenc}
\usepackage{pifont}
\usepackage{graphicx} 
\usepackage{subcaption}
\usepackage{wrapfig}
\usepackage{multirow}
\usepackage{booktabs}
\usepackage{xcolor}
\usepackage{color}       
\usepackage{soul}
\usepackage{mathtools}
\usepackage{mathrsfs}

\usepackage{colortbl}
\usepackage[OT1]{fontenc}
\usepackage[misc]{ifsym}
\usepackage{amssymb}

\definecolor{bg}{rgb}{0.8, 0.5, 0.8}
\definecolor{tone}{rgb}{0.9, 0.5, 0.2}
\definecolor{edit}{rgb}{0.4, 0.7, 0.2}
\definecolor{id}{rgb}{0.3, 0.6, 0.9}
\definecolor{airforceblue}{rgb}{0.6, 0.7, 0.8}
\definecolor{babyblue}{rgb}{0.7, 0.8, 0.9}

\newcommand{\customfootnotetext}[1]{%
  \begingroup
    \renewcommand{\thefootnote}{}
    \footnotetext{#1}%
  \endgroup
}

\title{\textbf{DisEnvisioner}:~\underline{Dis}entangled and~\underline{En}riched~\underline{Visual} Prompt for Customized Image Generation}

\author{\textbf{Jing He\textsuperscript{1\textcolor{red}{\ding{81}}}}
\textbf{Haodong Li\textsuperscript{1\textcolor{red}{\ding{81}}}}
\textbf{Yongzhe Hu\textsuperscript{\empty}}
\textbf{Guibao Shen\textsuperscript{1}}
\textbf{Yingjie Cai\textsuperscript{3}}
\textbf{Weichao Qiu\textsuperscript{3}}
\textbf{Yingcong Chen\textsuperscript{1,2 \Letter}}\\
$^{1}$HKUST(GZ) $^{2}$HKUST
$^{3}$Noah's Ark Lab\\
{
\fontfamily{cmtt}\selectfont\{jhe812, hli736\}@connect.hkust-gz.edu.cn;  yingcongchen@ust.hk
} 
}

\newcommand{\haodong}[1]{#1}
\newcommand{\jing}[1]{#1}

\iclrfinalcopy 
\begin{document}

\maketitle

\vspace{-11mm}
\begin{figure}[h]
    \centering
    \includegraphics[width = 0.98\linewidth]{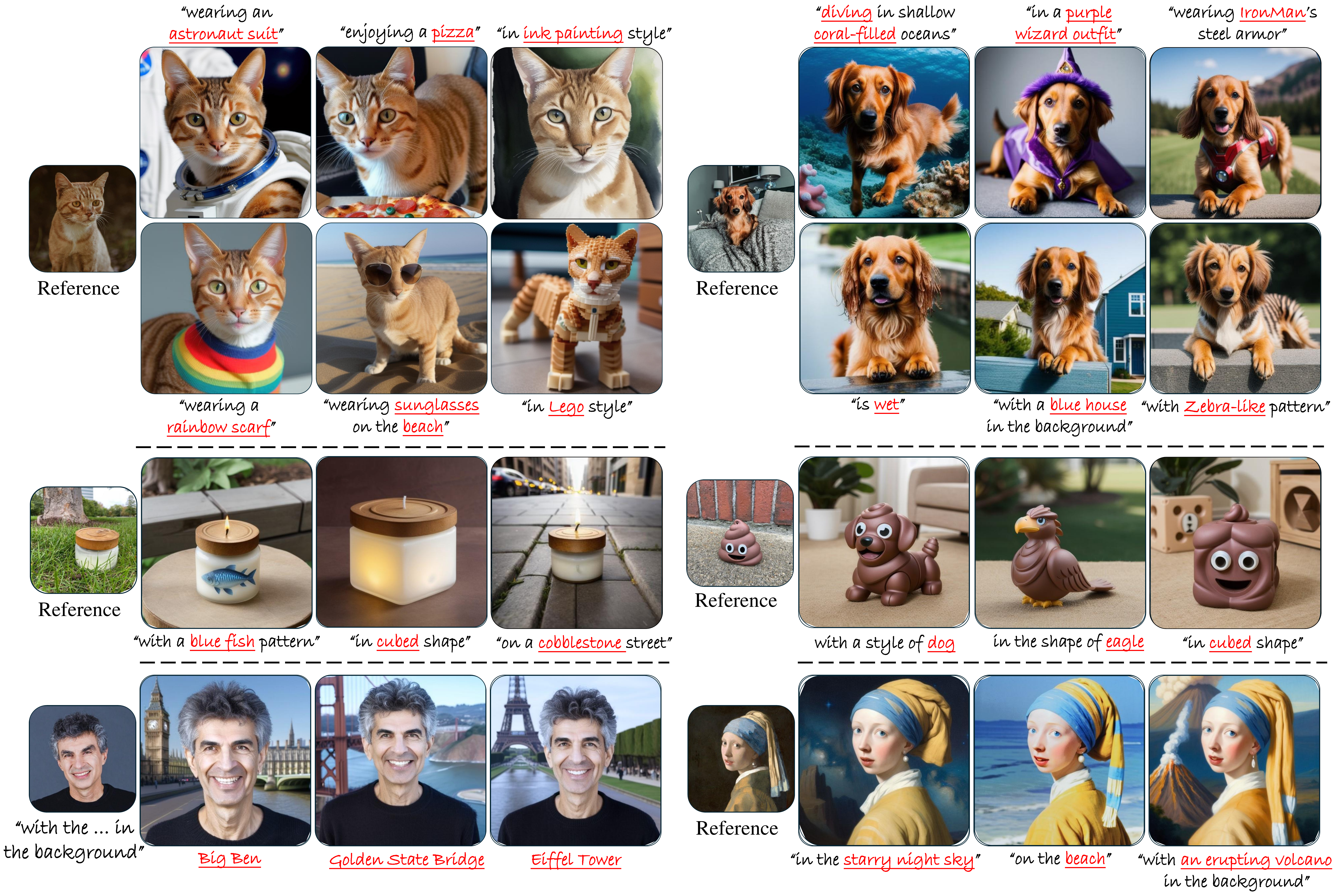}
    \caption{\textbf{Customization examples of DisEnvisioner.} Without cumbersome tuning or relying on multiple reference images, DisEnvisioner is capable of generating a variety of exceptional customized images. Characterized by its emphasis on the interpretation of subject-essential attributes, DisEnvisioner effectively discerns and enhances the subject-essential feature while filtering out irrelevant attributes, achieving superior personalizing quality in both editability and ID consistency.}
    \label{fig:teaser}
\end{figure}

\customfootnotetext{\textsuperscript{\textcolor{red}{\ding{81}}}Both authors contributed equally. \textsuperscript{\Letter} Corresponding author.}


\vspace{-1mm}
\begin{abstract}
\vspace{-1mm}
In the realm of image generation, creating customized images from visual prompt with additional textual instruction emerges as a promising endeavor. However, existing methods, both tuning-based and tuning-free, struggle with interpreting the subject-essential attributes from the visual prompt. This leads to subject-irrelevant attributes infiltrating the generation process, ultimately compromising the personalization quality in both editability and ID preservation. In this paper, we present \textbf{DisEnvisioner}, a novel approach for effectively extracting and enriching the subject-essential features while filtering out -irrelevant information, enabling exceptional customization performance, in a \textbf{tuning-free} manner and using only \textbf{a single image}. Specifically, the feature of the subject and other irrelevant components are effectively separated into distinctive visual tokens, enabling a much more accurate customization. Aiming to further improving the ID consistency, we enrich the disentangled features, sculpting them into a more granular representation. Experiments demonstrate the superiority of our approach over existing methods in instruction response (editability), ID consistency, inference speed, and the overall image quality, highlighting the effectiveness and efficiency of DisEnvisioner.
Project page: \href{https://disenvisioner.github.io/}{\textcolor{blue}{\fontfamily{cmtt}\selectfont{disenvisioner.github.io}}}.
\end{abstract}

\section{Introduction}
\label{sec:intro}
By training with billions of image-text pairs, state-of-the-art text-to-image generation models, \textit{e.g.}, DALL$\cdot$E~\citep{dalle}, Imagen~\citep{imagen}, UnCLIP~\citep{unclip}, Stable Diffusion (SD)~\citep{stablediffusion}, and PixArt-$\alpha$~\citep{pixart}, have demonstrated remarkable proficiency in generating contextually aligned images from textual descriptions.
Despite unprecedented creative capabilities of these text-to-image models, customized image generation poses a new and more intricate challenge. This task aims to synthesize life-like imagery that not only accurately responds to natural language instructions (\textbf{editability}) but also preserves the subject's identity based on reference images (\textbf{ID consistency}), is garnering great attention from both academia and industry~\citep{disenbooth, dreamartist, instantbooth, subjectdiffusion, TI, customdiffusion, dreambooth, blipdiffusion, elite, ip-adapter, photomaker, instantid, photoverse, arar2024palp, voynov2023p+}.

\begin{figure}[!t]
    \centering
    \includegraphics[width=\textwidth]{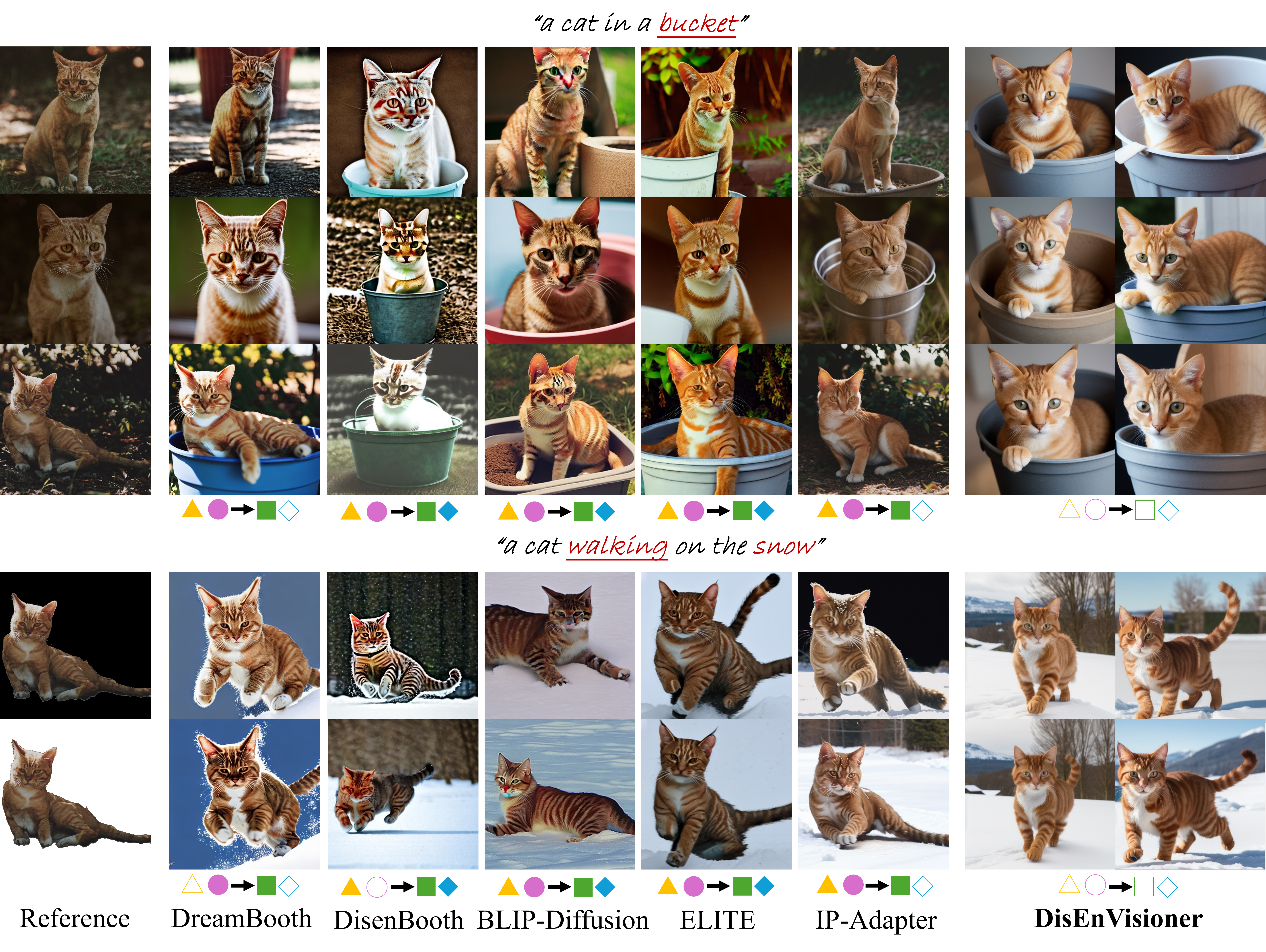}
    \caption{
    \textbf{Comparisons between DisEnvisioner and existing methods}~\citep{dreambooth, disenbooth, blipdiffusion, elite, ip-adapter} \textbf{under single-image setting}.
    We evaluate these methods on the same subject with different poses and environments. 
    It can be observed that irrelevant factors, such as the \sethlcolor{yellow}\hl{subject's posture} (\includegraphics[width=8pt]{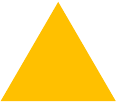}) and \sethlcolor{bg}\hl{background} (\includegraphics[width=8pt]{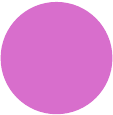}),
    can affect the customization quality and result in \sethlcolor{edit}\hl{poor editability} (\includegraphics[width=8pt]{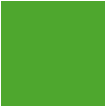}) or \sethlcolor{id}\hl{poor ID consistency} (\includegraphics[width=8pt]{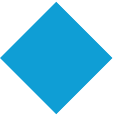}). 
    For instance, BLIP-Diffusion falls in  both two factors, leading to poor editability and ID consistency. We denotes its performance as ``\includegraphics[height=8pt]{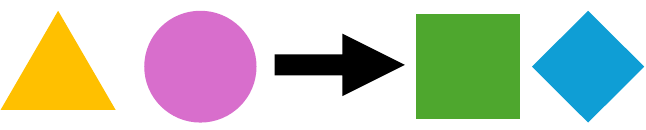}'' \haodong{(the symbols without color filling, such as ``\includegraphics[height=8pt]{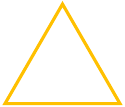}'', indicates that the customization is \textit{not} affected by subject’s posture, and ``\includegraphics[height=8pt]{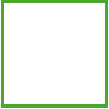}'' indicates \textit{good} editability)}.
    We \haodong{also} try to use masks to filter out irrelevant information for these methods. However, the harmful influence of subject's posture still exists. And solid background colors (\emph{e.g.}, white or black) also can \haodong{harmfully impact the} customization quality\haodong{, leading to textureless backgrounds}.
    } 
    \label{fig:motivation}
\end{figure}

For high-quality image customization, accurate interpretation of the visual prompt (\textit{i.e.}, the input image) is crucial. This involves effectively extracting \textbf{subject-essential} attributes from the reference image while minimizing the influence of \textbf{subject-irrelevant} attributes.
Failure to do so may result in
1) overemphasis on irrelevant details: generated images may prioritize irrelevant information, sidelining the textual instructions and compromising the overall editability;
2) diminished subject identity: the feature of subject-essential attributes becomes entangled with irrelevant information, degrading the subject representation.

Existing methods, both tuning-based~\citep{dreambooth, TI, customdiffusion, disenbooth,dreamartist} and tuning-free~\citep{elite, blipdiffusion, ip-adapter,subjectdiffusion,photomaker}, struggle to accurately interpret subject-essential attributes, particularly given only a single reference image. 
Specifically, prevailing tuning-based methods
like DreamBooth~\citep{dreambooth} and DisenBooth~\citep{disenbooth}, heavily rely on multiple reference images to capture the common subject concept into the model. 
Despite their impressive results, it is difficult to interpret the subject-essential attributes in single-image scenarios, leading to compromised customization quality (Fig.~\ref{fig:motivation}). 
Additionally, each subject requires individual fine-tuning, which is time-consuming and hinders practical application.
Recent tuning-free methods~\citep{elite, subjectdiffusion, photomaker, instantbooth,ip-adapter,blipdiffusion, photoverse}, aim to offer a significant boost in inference speed, and can generate customized images with a single reference image. 
However, these approaches still fail to accurately disentangle the subject-essential features from the reference image.
Among those methods, IP-adapter~\citep{ip-adapter} and PhotoVerse~\citep{photoverse} directly treat the \textit{global} feature of the given image as \textit{subject-essential}, inevitably introducing the subject-irrelevant information that diminishes the personalizing quality. 
Moreover, ELITE~\citep{elite} and BLIP-Diffusion~\citep{blipdiffusion} attempt to learn subject representations, but their implicit extraction of subject features \haodong{proves ineffective (please see Sec.~\ref{sec:re_cig} for detailed discussion)}, \jing{also} leading to unsatisfactory customization performance hindered by irrelevant information.
As illustrated in Fig.~\ref{fig:motivation}, the customization quality of these methods are notably compromised by irrelevant factors (we conclude the primary irrelevant factors in this case as: subject's posture, background and image tone). 
A direct solution of these methods to prevent the irrelevant information is to use a segmentation mask to remove the background, but it is inadequate. As shown in Fig.~\ref{fig:motivation}: \haodong{\ding{172} factors like the subject's pose still introduce irrelevant details; \ding{173} replacing the background with solid colors (\emph{e.g.}, white or black) also influence the customization, potential leading to texture-less backgrounds in the generated customized images.}
Thus, an effective disentanglement of the subject-essential features is indeed necessary.


\begin{figure}[!t]
    \centering
    \includegraphics[width=0.75\textwidth]{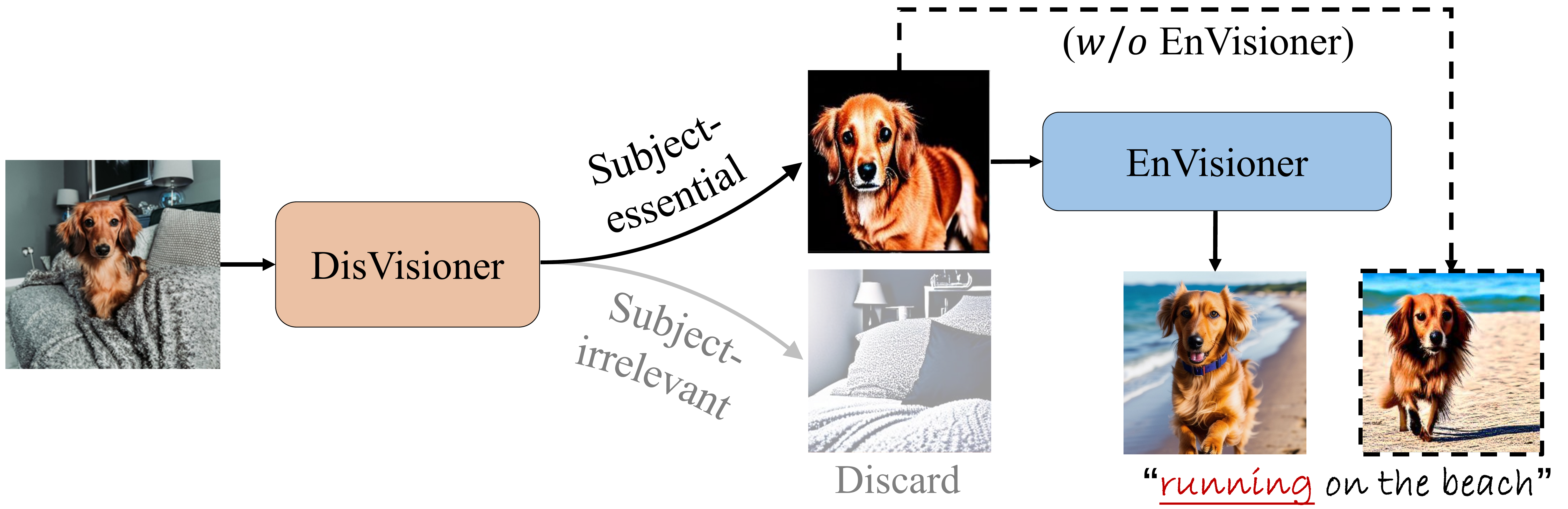}
    \caption{\textbf{Overview of DisEnvisioner}, which consists of two key components: \textbf{DisVisioner} and \textbf{EnVisioner}. During inference, subject-irrelevant features are discarded to avoid harmful disturbances.}
    \label{fig:method_overview}
\end{figure}
Motivated by the above analysis, 
we propose \textbf{DisEnvisioner}, a novel framework meticulously designed to addresses the core issues via feature disentanglement and enrichment. 
As illustrated in Fig.~\ref{fig:method_overview}, the image is tokenized into compact disentangled features, \emph{i.e.}, subject-essential and subject-irrelevant tokens, through the DisVisioner.
Thus, the subject-irrelevant features can be filtered, making
sure the model only focus on essential attributes of the subject, facilitating more accurate editability during customization.
The disentangled subject-essential features are further enriched by EnVisioner before feeding into the pre-trained SD model~\citep{stablediffusion} for customized generation, significantly boosting the ID-consistency and the overall customization quality. Together with above innovations, we achieve both accurate and high-quality image customization.

In summary, our key contributions are as follows.
\begin{itemize}

\item We emphasize the critical role of subject-essential attribute in customized image generation, which is the foundation of faithful subject concept reconstruction and reliable editability, thereby ensuring more accurate customization under more diverse textual instructions.
\item We present DisEnvisioner, a simple yet effective framework designed for single-image, tuning-free image customization, utilizing visual disentanglement and enrichment.
\item Comprehensive experiments validate DisEnvisioner's superiority in adhering to instructions, maintaining ID consistency, and inference speed, demonstrating its superior personalization capabilities and efficiency.
\end{itemize}

\section{Related Works}
\label{sec:rw}
\subsection{Text-to-Image Generation}

In the field of text-to-image generation, the evolution of methodologies has transitioned from Generative Adversarial Networks (GANs)~\citep{goodfellow2014generative, zhang2017stackgan, zhang2018stackgan++, zhang2021cross, pixelfolder, StyleGAN1, StyleGAN2, StyleGAN3} to advanced Diffusion Models~\citep{ho2020denoising,unclip,imagen,dalle,nichol2021glide,pixart,he2024lotus, stablediffusion}.
Early GAN-based models like StackGAN~\citep{zhang2017stackgan, zhang2018stackgan++}, AttnGAN~\citep{xu2018attngan}, and XMC-GAN~\citep{zhang2021cross} introduced multi-stage generation and attention mechanisms to improve image fidelity and alignment with textual descriptions. 
A significant breakthrough, DALL$\cdot$E~\citep{dalle}, utilizes a transformer and auto-regressive model to merge text and image data, trained on 250M pairs for high-quality, intuitive image synthesis. 
Following this, a series of diffusion-based methods such as GLIDE~\citep{nichol2021glide}, DALL$\cdot$E2~\citep{unclip}, and Imagen~\citep{imagen} have been introduced, offering enhanced image quality and textual coherence. 
The Latent Diffusion Model (LDM)~\citep{stablediffusion},
trained on 5 billion pairs, further enhanced training efficiency without compromising performance, becoming a community standard. 
Despite remarkable strides in generating images from textual descriptions, current methodologies fall short in rendering customized visual concepts from reference images. In our paper, we dive into customized image generation, extending the capabilities of existing text-to-image techniques to craft personalized visual concepts.

\subsection{Customized Image Generation}
\label{sec:re_cig}
Existing methods in the field of customized image generation primarily fall into two categories: tuning-based and tuning-free.
Tuning-based methods~\citep{TI,dreambooth,dreamartist,svdiff,customdiffusion,disenbooth,hua2023dreamtuner} involve fine-tuning a pre-trained generative model with several reference images of a specific concept during test-time. 
Despite the effectiveness of these methods, their high computational demand and scalability challenges limit the practicality.
In contrast, tuning-free methods~\citep{tamingencoder,instantbooth,blipdiffusion,subjectdiffusion,ip-adapter,elite,photoverse,photomaker,instantid}
employ advanced encoders to represent customized visual concepts, enabling image generation without the need of test-time fine-tuning and leading to significantly improved efficiency. 
However, InstantBooth~\citep{instantbooth} and PhotoMaker~\citep{photomaker} still rely on multiple reference images to ensure the ID fidelity, and more recent works start to focus on singe-image scenarios.  
IP-Adapter~\citep{ip-adapter} introduces a coupled cross-attention mechanism that handles visual and textual prompts separately, allowing for fine-grained visual concept generation using only a single reference image.  
PhotoVerse~\citep{photoverse} employs a similar dual-attention strategy, with an additional identity loss to further enhance ID consistency. However, both IP-Adapter and PhotoVerse directly treat the global feature of the given image as subject-essential, inevitably introducing the subject-irrelevant information that diminishes the personalizing quality.
BLIP-Diffusion~\citep{blipdiffusion} and ELITE~\citep{elite} attempt to learn subject-essential representations. BLIP-Diffusion derives subject representations by querying image features with subject names using BLIP-2~\citep{li2023blip}.
\jing{However BLIP-2 is pre-trained to extract \textit{global} image feature aligned with \textit{global} text prompts. It may struggle to accurately query clean \textit{local} subject-essential features when the given subject names describe only the \textit{local} information of a whole image. Thus, BLIP-Diffusion may still fall short in filtering out subject-irrelevant attributes.}
%
ELITE~\citep{elite} separates subject attributes from others by utilizing CLIP features from different layers, assuming that the subject is represented by the deeper features and other concepts by shallower features. However, since the deep and shallow features are not fully disentangled, the subject attributes obtained through this multi-layer approach still remains inaccurate.
Additionally, ELITE requires segmentation masks to combine the masked CLIP image features and the obtained subject feature for enhanced details. However, the segmenting process is cumbersome and the masked image feature is still inaccurate and can introduce subject-irrelevant factors.
%
\jing{In this paper, although DisEnvisioner also adopts a two-stage pipeline following BLIP-Diffusion and ELITE, unlike these methods, 
DisEnvisioner focuses more on explicitly extracting the subject-essential attributes via DisVionser (Sec.~\ref{subsec:disvisioner}), achieving more accurate customization. Then, we enrich the disentangled features by EnVisioner (Sec.~\ref{subsec:envisioner}), rather than introducing additional features as in ELITE~\citep{elite}, to further enhance the ID consistency without unnecessary disturbance.   
}

\section{Method}
\label{sec:method}
The goal of customized image generation is to synthesize lifelike images that adhere to the textual instructions while preserving the subject's identity from the reference image. 
To tackle the pivotal challenge of minimizing the influence of subject-irrelevant attributes and enhancing subject-essential attributes, we introduce DisEnvisioner—a novel approach focused on disentangling and enriching subject-essential attributes.
As depicted in Fig.~\ref{fig:method_overview}, DisEnvisioner initiates the process by disentangling the image features into subject-essential and -irrelevant attributes by the DisVisioner. Subsequently, to bolster the consistency of the subject's identity, the EnVisioner is further employed to refine the subject-essential features into a more fine-grained representation. Our discussion commences with an overview of basic concepts (Sec.~\ref{subsec:pre}), followed by an in-depth exploration of the DisVisioner (Sec.~\ref{subsec:disvisioner}) and EnVisioner (Sec.~\ref{subsec:envisioner}). 

\subsection{Preliminaries}
\label{subsec:pre}
Diffusion models~\citep{ho2020denoising,unclip,imagen,dalle,nichol2021glide,pixart} represent the current state-of-the-art in generative modeling for high-fidelity image generation, it involves forward diffusion and reverse denoising processes. The forward process incrementally introduces noise to the data, transforming it into a Gaussian random noise by a Markov chain over a fixed number of steps $T$. In the reverse phase, a learned neural network is utilized to predict the added noise at each step, thereby recovering the original data from the noisy sample. This noising-denoising mechanism enables diffusion models to achieve impressive results in generating high-quality images with fine details and realism.

In this study, we adopt Stable Diffusion (SD)~\citep{stablediffusion}, a latent diffusion model built upon UNet~\citep{ronneberger2015u}, as our foundational generative model.
Firstly, an auto-encoder (VAE) $\left\{\mathscr{E}(\cdot), \mathscr{D}(\cdot)\right\}$ is trained to map between RGB space and the latent space, \textit{i.e.}, $ \mathscr{E}(\mathbf{x}) = \mathbf{z} $, $\mathscr{D}(\mathscr{E}(\mathbf{x}))\approx \mathbf{x} $. And the textual conditions are obtained using a pre-trained CLIP text encoder $\boldsymbol{c} = \psi_\theta^{\text{T}}(y)$, where  $y$  is the given prompt.
The training objective is:
\begin{equation}
\label{eq:duffusion_loss}
   L_{\text{LDM}}(\theta) = \mathbb{E}_{\textbf{x}_0,\boldsymbol{c},\boldsymbol{\epsilon}_t\sim\mathcal{N}(0,1),t\in[1,T]}\Big[\|\boldsymbol{\epsilon}_t-\boldsymbol{\epsilon}_\theta(\textbf{z}_t,\boldsymbol{c},t)\|_2^2\Big],
\end{equation} 
where $\boldsymbol{\epsilon}_t$ represents the noise introduced during the forward process, and $\boldsymbol{\epsilon}_\theta(\textbf{z}_t,\boldsymbol{c},t)$ is the predicted noise. 
Cross-attention is adopted to introduce textual condition into SD.
Specifically, the latent image feature $\boldsymbol f\in \mathbb{R}^{(\text{H}\times\text{W})\times d_{\text{q}}}$ and text condition $\boldsymbol{c}\in\mathbb{R}^{n_y\times  d_{\text{k}}}$ are firstly projected to obtain query $\boldsymbol  Q = \boldsymbol w_{\texttt{to\_q}}\circ \boldsymbol f $, key $\boldsymbol  K =\boldsymbol w_{\texttt{to\_k}} \circ \boldsymbol{c}  $, and value $\boldsymbol V = \boldsymbol w_{\texttt{to\_v}} \circ \boldsymbol{c} $, where $\boldsymbol w$ are weights of the corresponding mapping layers.
Then, the cross-attention is calculated as:
\begin{equation}
    \label{eq:crossattn}
    \operatorname{Attention}(\boldsymbol Q,\boldsymbol  K, \boldsymbol V)=\operatorname{SoftMax}\left(\frac{\boldsymbol Q \boldsymbol K^\text{T}}{\sqrt{d_{\text{k}}}}\right) \boldsymbol  V .
\end{equation}
During inference, the model iteratively constructs images $ \mathbf{x}_0 $ from random noises $\mathbf{z}_T$ through reverse denoising process.

\begin{figure}[!t]
\centering
\subcaptionbox{\textbf{DisVisioner} \label{subfig:vd}}{
\includegraphics[width = 0.38\linewidth]{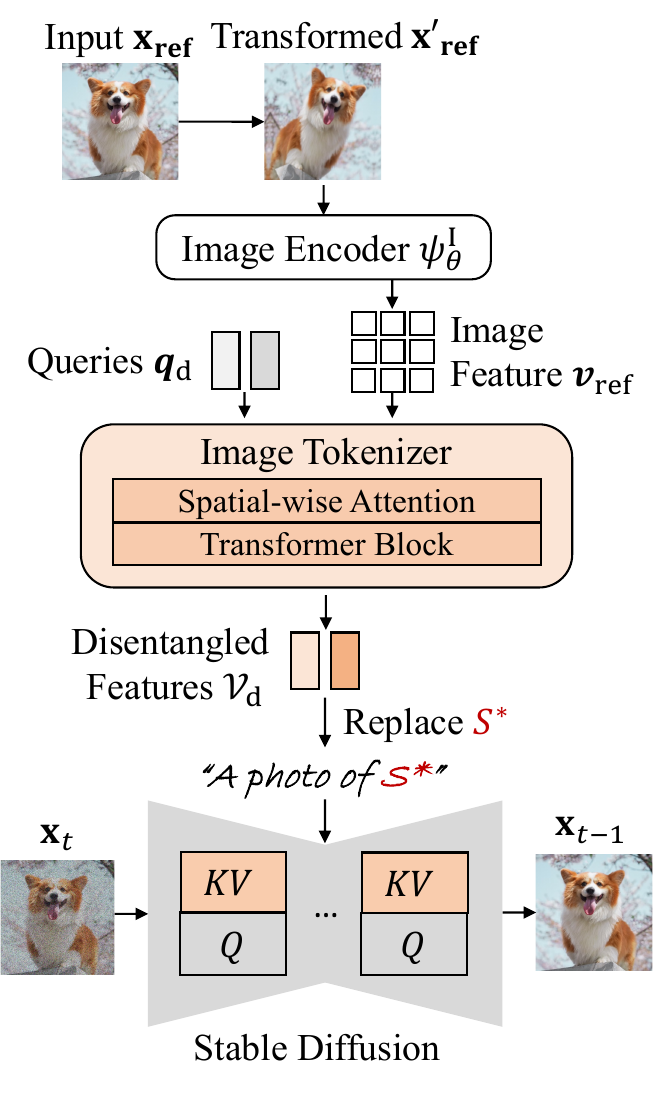}
}
\hfill
\subcaptionbox{\textbf{EnVisioner} \label{subfig:ve}}{
\includegraphics[width = 0.58\linewidth]{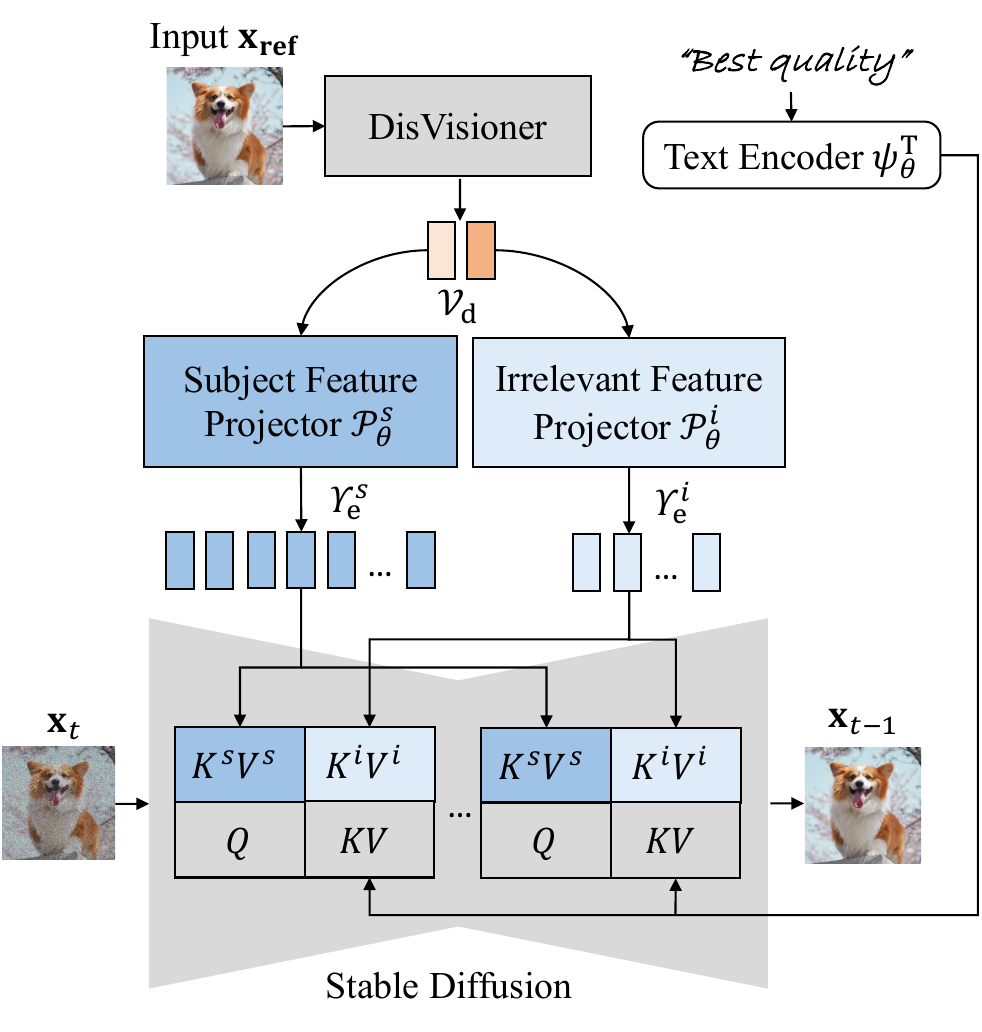}}
\caption{
\textbf{Training pipeline of DisEnvisioner.} Our approach is structured into two stages: (a) DisVisioner firstly disentangles the features of the subject and other irrelevant components by aggregating the image feature $\boldsymbol{\upsilon}_{\text{ref}}$ into two distinct and orthogonal tokens $\boldsymbol{v}_{\text{d}}$. 
(b) EnVisioner subsequently refines and sculpts the disentangled features $\mathcal V_{\text{d}}$ into more granular representations to produce high ID-consistency images with the input image $\mathbf{x}_{\text{ref}}$, and can improve the overall visual quality of the image. Only colored modules (\textcolor{orange}{orange} and \textcolor{airforceblue}{blue}) are trainable.}
\label{fig:structure}
\end{figure}

\subsection{DisVisioner}
\label{subsec:disvisioner}
As revealed in~\citep{elite,blipdiffusion,ip-adapter}, the customized image can be encoded as a sequence of tokens via inversion. In our work, we propose to disentangle the image feature into subject-essential and -irrelevant tokens using an image tokenizer~\citep{wu2022learning}.
Image tokenizer is a method to aggregate the image features into compact visual tokens, with each token corresponds to a distinct visual component. As illustrated in Fig.~\ref{subfig:vd}, given a reference image $\mathbf{x}_{\text{ref}}$, we firstly transform them using an augmentation set. The transformed image $\mathbf{x'}_{\text{ref}}$ ensures that the model can extract the effective subject-essential feature to reconstruct the original image, rather than merely duplicating the subject.  Subsenquently, we employ the CLIP image encoder to extract image features $ \boldsymbol{\upsilon}_{\text{ref}} = \psi_\theta^{\text{I}}(\mathbf{x'}_{\text{ref}}) \in \mathbb{R}^{(\text{H}\times\text{W})\times d_{\text{k}}}$. 
The image tokenizer network $M(\cdot)$ is then utilized to obtain the disentangled visual tokens $\mathcal{V}_{\text{d}}$:
\begin{equation}
    \label{eq:tokenizer}
    \mathcal{V}_{\text{d}} = M (\boldsymbol{q}_{\text{d}}, \psi_\theta^{\text{I}}(\mathbf{x'}_{\text{ref}}))
\end{equation}
where  $ \boldsymbol{q}_{\text{d}} \in\mathbb{R}^{(n_{{s}} + n_{{i}})\times d_{\text{q}}} $ is the queries
for feature aggregation, $n_s$ for the subject feature while $n_{{i}}$ for other irrelevant features. In image tokenizer $M(\cdot)$, a spatial-wise cross-attention mechanism is firstly adopted to aggregate the image features, where $ \boldsymbol{q}_{\text{d}}$ is the query and $\psi_\theta^{\text{I}}(\mathbf{x}_{\text{ref}})$ serves as the key and value. The image features are separated into two mutually independent and orthogonal sets of features according to $ \boldsymbol{q}_{\text{d}}$ after the spatial-wise attention mechanism. \haodong{This mutual independence and orthogonality are ensured by the $\operatorname{SoftMax}(\cdot)$ function applied at spatial dimension within the spatial-wise attention.}
In order to determine the sequence of disentangled features, we initialize $\boldsymbol{q}_{\text{d}}$ with the CLIP prior, as revealed in~\citep{li2023clip}. Specificallt, the $n_s$ queries for subject are initialized with random vectors, while the remaining $n_i$ queries are initialized using CLIP class name embeddings to query the subject-irrelevant tokens. Transformer blocks are followed to further refine the disentangled features to finally obtain $\mathcal{V}_{\text{d}}$ as the output of our DisVisioner.

To train the image tokenizer $M(\cdot)$, we insert the disentangled features $\mathcal{V}_{\text{d}}$ into textual feature space of the prompt by replacing the textual embedding of placeholder $S^*$, and adopt the Eq.~\ref{eq:duffusion_loss} as the training objective for the target of reconstruction. Following~\citep{elite,customdiffusion},  the entire SD model is frozen except the \texttt{to\_k} and \texttt{to\_v} layers of the cross-attention module for correct interpretation of the new disentangled tokens. In our implementation, we set $n_s=1$ and $n_i=1$ for subject-essential and irrelevant features, respectively. Excessive tokens will lead to inaccurate disentanglement, thus impairing the customization quality (please refer to Sec.~\ref{subsec:ablation} for more details). 

Benefiting from image tokenizer, the subject-essential features are accurately compressed into the tokens, clearly and accurately separated from irrelevant features via the spatial-wise attention space. During customized image generation, the subject-irrelevant token will be discarded for exclude the unwanted disturbance, facilitating the editing accuracy and ID fidelity. 

\subsection{EnVisioner}
\label{subsec:envisioner}
While DisVisioner effectively extracts the subject-essential feature into a single token, it may be inadequate for capturing the detailed nuances of the customized subject. In EnVisioner, we
map the disentangled features into a sequence of granular tokens using new projectors $P^s(\cdot)$ and $P^i(\cdot)$. The utilization of separate projectors also guarantees the disentanglement between the subject-essential and -irrelevant tokens.

As shown in Fig.~\ref{subfig:ve}, the disentangled tokens $\mathcal{V}_{\text{d}}$ from DisVisioner is enriched into multiple tokens:
\begin{equation}
    \label{eq:proj}
    \Upsilon^s_\text{e} =   P^{{s}} \circ \boldsymbol  \tau^{{s}}_{\text{d}},\ \Upsilon^i_\text{e} =  P^{{i}} \circ \boldsymbol  \tau^{{i}}_{\text{d}},
\end{equation}
where $\tau^{{s}}_{\text{d}}, \tau^{{i}}_{\text{d}} \in \mathbb{R}^{1\times d}$ denote the disentangled subject-essential and the irrelevant feature through DisVisioner ($\mathcal{V}_{\text{d}}=[\tau^{{s}}_{\text{d}}, \tau^{{i}}_{\text{d}}]$). $\Upsilon^s_\text{e} \in \mathbb{R}^{n^{\prime}_s \times \text{d}}$ is the enriched subject-essential tokens from $\tau^{{s}}_{\text{d}}$, and $\Upsilon^i_\text{e} \in \mathbb{R}^{n^{\prime}_i \times \text{d}}$ is the irrelevant tokens projected from $\tau^{{i}}_{\text{d}}$.
By enriching the subject-essential token into multiple ones, the disentangled concept is further enhanced into a more granular representation, improving especially the ID consistency between the synthesized image and the reference image.

Similar to DisVisioner, we also employ Eq.~\ref{eq:duffusion_loss} as the training objective to train the feature projectors in EnVisioner, while keeping the entire SD model and all DisVisioner modules frozen. Additionally, we separatelt introduce cross-attention layers for subject-essential tokens and -irrelevant ones. This separate injection strategy further ensures that the subject feature will not be interfered with by other irrelevant factors.
Specifically, given the latent diffusion feature $\boldsymbol f\in \mathbb{R}^{(\text{H}\times\text{W})\times d_{\text{q}}}$ and text instruction $\boldsymbol{c}\in\mathbb{R}^{n_y\times  d_{\text{k}}}$, the cross-attention output $\boldsymbol f^{\prime}$ is derived through three decoupled cross-attention layers, which can be described via the following Equation:

\begin{equation}
    \label{eq:envisioner}
    \small
    \begin{aligned}
    \boldsymbol f^{\prime} = \operatorname{Attention}(\boldsymbol Q, \boldsymbol K, \boldsymbol V) + \lambda_s\operatorname{Attention}(\boldsymbol Q, \boldsymbol K^s, \boldsymbol V^s) + \lambda_i\operatorname{Attention}(\boldsymbol Q, \boldsymbol K^i, \boldsymbol V^i),\\
    \text{where } K^s = \boldsymbol w_{\texttt{to\_k}}^{s}\circ \Upsilon^s_\text{e}, V^s = \boldsymbol w_{\texttt{to\_v}}^{s}\circ \Upsilon^s_\text{e};  K^i = \boldsymbol w_{\texttt{to\_k}}^{i}\circ \Upsilon^i_\text{e}, V^i = \boldsymbol w_{\texttt{to\_v}}^{i}\circ \Upsilon^i_\text{e},  
    \end{aligned}
\end{equation}

the $\operatorname{Attention}(\cdot)$ is defined in Eq.~\ref{eq:crossattn}, $\boldsymbol w_{\texttt{to\_k}}^{s}$, $\boldsymbol w_{\texttt{to\_v}}^{s}$, $\boldsymbol w_{\texttt{to\_k}}^{i}$ and $\boldsymbol w_{\texttt{to\_v}}^{i}$ are trainable mapping layers. During training, weights $\lambda_s$ and $\lambda_i$ are fixed at $1.0$. During inference, for the purpose of effectively ignoring subject-irrelevant features encoded in $\Upsilon^i_\text{e}$ , we set $\lambda_i=0$.

\section{Experiments}
\label{sec:exp}
\subsection{Experimental Setup}
\subsubsection{Training Dataset}
We use the \textit{training set} of OpenImages V6~\citep{kuznetsova2020open} to train the DisEnvisioner. 
It contains about 14.61M annotated boxes across 1.74M images. 
Based on this dataset, we construct 6.82M \{prompt, image\} pairs for training, where the images are cropped and resized ($256\times 256$ for DisVisioner and $512\times 512$ for EnVisioner) according to the bounding box annotations, and the text prompts are obtained by randomly selecting a CLIP ImageNet template~\citep{radford2021learning} and integrating the annotated class names into it. 

\subsubsection{Evaluation Dataset and Metrics}
\label{sec:metrics}
The evaluation is carried out on the DreamBooth~\citep{dreambooth} dataset, which comprises 30 subjects and 158 images in total (4$\sim$6 images per subject). 
For quantitative evaluation, 25 editing prompts~\citep{dreambooth} are used for each image. We inference 40 times for each \{prompt, image\} pair, generating 158,000 customized images for \haodong{evaluation across 6 metrics}.

In alignment with previous methodologies~\citep{dreambooth, customdiffusion, elite, photomaker}, we utilize: 
1) CLIP Text-alignment (\textbf{C-T}) to measure the instruction response fidelity; 2) CLIP Image-alignment (\textbf{C-I}) and 3) DINO Image-alignment (\textbf{D-I})~\citep{dino} to evaluate the ID consistency with the reference image. 
Additionally, we introduce a novel metric, \emph{i.e.}, Internal Variance (\textbf{IV}) of image-alignment, aiming at quantifying the impact of irrelevant factors. IV measures the variance of customization results for images containing the same subject but different environments (like the top three rows in Fig.~\ref{fig:motivation}). Lower IV value indicates that the results only be controlled by the subject and textual instruction, rather than the irrelevant factors from environments.
In terms of efficiency, we record the 6) Inference-time (\textbf{T}) on single NVIDIA A800 GPU for evaluation. Moreover, we calculate the mean ranking (\textbf{mRank}) of all metrics for each method to show the comprehensive performance. 

\subsubsection{Implementation Details.}
\label{subsec:impl}
DisEnvisioner is built upon Stable Diffusion v1.5
, employing OpenCLIP ViT-H/14 
model as the image/text encoder. 
During training, DisVisioner is configured with batch size of 160, learning rate of 5e-7 at the resolution of 256. The training steps is 120K. We set the token number $n_s = 1 $ and $ n_i=1$,
for subject-essential feature and -irrelevant respectively. 
The EnVisioner employs the batch size of 40, learning rate of 1e-4 at the resolution of 512. The training steps is also 120K. The enriched token number is $n_s^{\prime} = 4$ and $n_i^{'} = 4$, with attention scale $\lambda_s = 1.0$ and $\lambda_i = 1.0$. 
All experiments are conducted on 8 NVIDIA A800 GPUs using the AdamW optimizer~\citep{loshchilov2017decoupled} with a weight decay of 0.01. 
To enable classifier-free guidance~\citep{dhariwal2021diffusion}, we use a probability of 0.05 to drop the condition, both textual and visual. 
During inference, $\lambda_i$ is set to 0 to eliminate irrelevant feature. In addition, we use the  DDIM sampler~\citep{song2020denoising} with 50 steps and the scale of classifier-free guidance is set to 5.0.

\begin{table}[!t]
\footnotesize
\centering
\caption{\textbf{Quantitative comparisons with existing methods}. The evaluation metrics include text alignment for assessing editability (C-T), image-alignment for ID-consistency (C-I, D-I), internal variance to demonstrate the resistance to subject-irrelevant factors (IV), and inference time for efficiency (T). DisEnvisioner demonstrates better comprehensive performance than other methods. Top results are in \textbf{bold}; second-best are \underline{underlined}.\textsuperscript{$*$}For equity, we assess only inference (and test-time tuning, if applicable) times, omitting I/O operations. $^\S$For equity, we consider C-I and D-I as two sub-indicators of image-alignment, the rank of each with a weight of \textbf{0.5} in the mRank calculation, while the ranks of all other metrics have a weight of \textbf{1.0}.
}

\setlength{\tabcolsep}{7.5pt}
\begin{tabular}{l|c|cc|c|c|c}
\toprule
Method & C-T$\uparrow$ & C-I$^\S$$\uparrow$ & D-I$^\S$$\uparrow$ 
& IV$\downarrow$ & T\textsuperscript{$*$}$\downarrow$ (s)\textsuperscript & \textbf{mRank}$\downarrow$ \\
\midrule
DisenBooth~\citep{disenbooth}  & 0.303 
& 0.760 & {0.781}  
& 0.041  & 2.42e+3&4.8 \\ 
DreamBooth~\citep{dreambooth}        & 0.286 
& \underline{0.842} & \underline{0.849} 
& 0.039  & 1.12e+3& 4.3\\ 
\midrule
ELITE~\citep{elite}             & 0.287 
& {0.792} & 0.770  
& 0.036  & 4.12& 4.1 \\
IP-Adapter~\citep{ip-adapter}      & 0.275 
& \textbf{0.883} & \textbf{0.912} 
& 0.033  & 1.98& {3.3}  \\
\underline{BLIP-Diffusion}~\citep{blipdiffusion}  & 0.295 
& 0.785 &  0.765 
& \underline{0.029}  & \textbf{1.10}& \underline{2.9} \\
\midrule
\textbf{DisEnvisioner}             &\textbf{0.315} 
&{0.828}  & {0.802} 
& \textbf{0.026} & \underline{1.96}& \textbf{2.0}\\
 \bottomrule
\end{tabular}
\label{tab:metrics}
\end{table}
%
%
\begin{figure}[t]
    \centering
    \includegraphics[width = 0.99\linewidth]{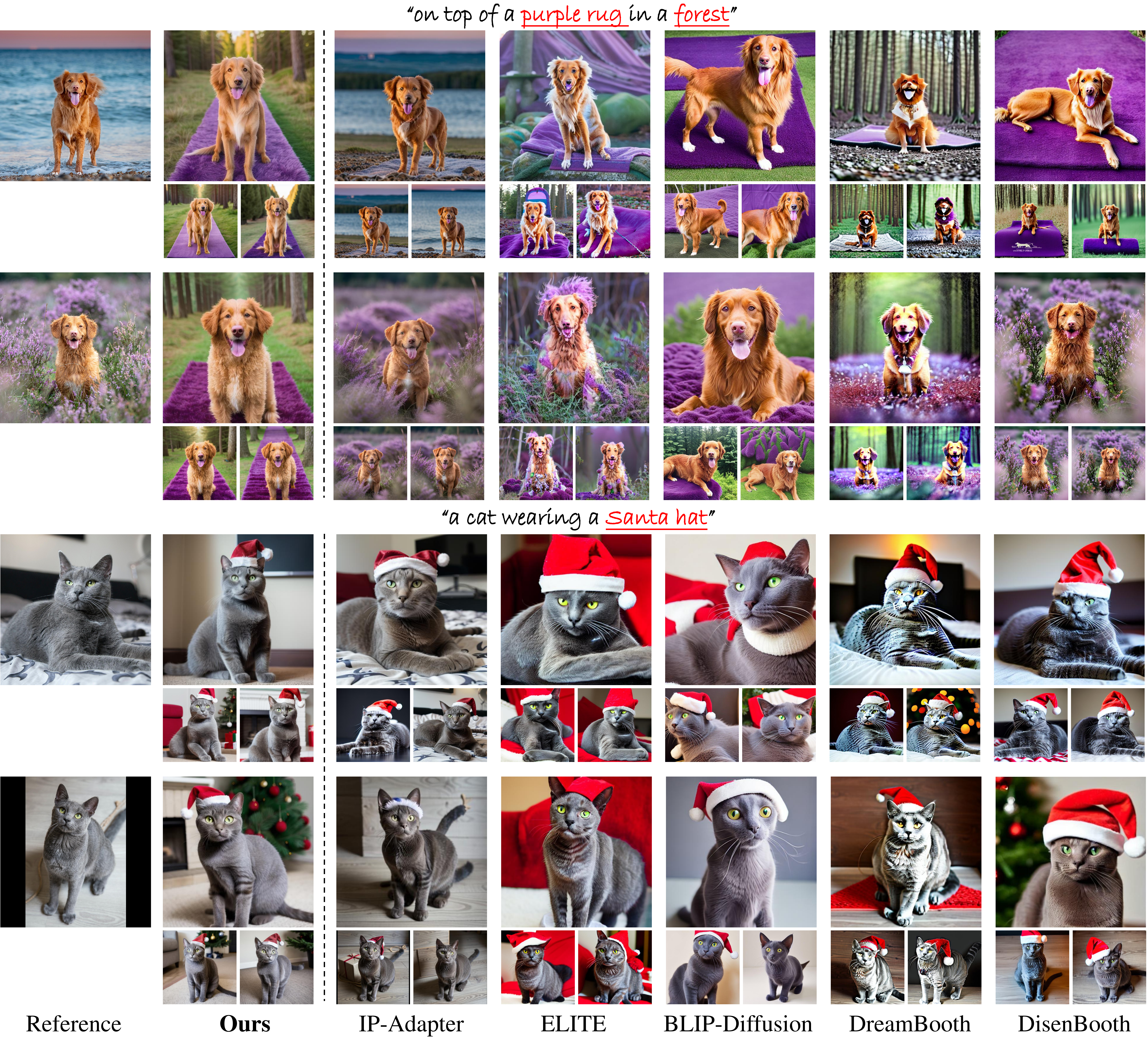}
    \caption{\textbf{Qualitative comparison on live subjects.}
    Comparing two subjects across various scenarios, DisEnvisioner excels in editability and ID-consistency. Notably, the animal's posture does not affect customization, showcasing our strength in capturing subject-essential features.}
    \label{fig:more_result1}
\end{figure}

\begin{figure}[t]
    \centering
    \includegraphics[width = 0.99\linewidth]{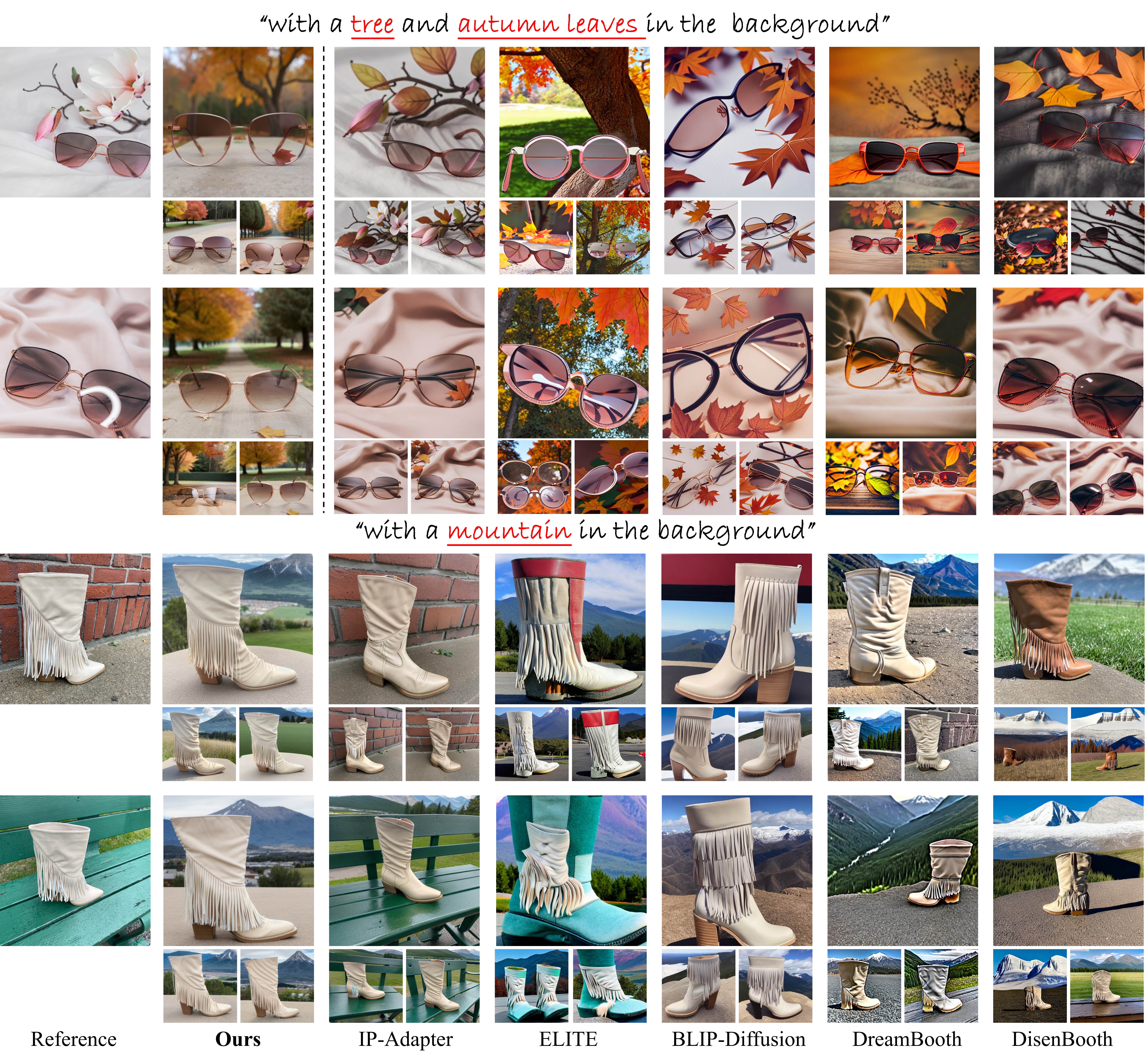}
    \caption{\textbf{Qualitative comparison on non-live subjects.}
    Besides offering better editability of textual instructions, DisEnvisioner also excels in preserving accurate subject identity. Moreover, images generated by DisEnvisioner are minimally affected by irrelevant elements of reference images.
    }
    \label{fig:more_result2}
\end{figure}

\subsection{Quantitative Results}
\label{subsec: quantitative}
We compare DisEnvisioner against five leading methods according to metrics specified in Sec.~\ref{sec:metrics}. 
For the tuning-based methods, we chose DreamBooth~\citep{dreambooth} and DisenBooth~\citep{disenbooth}, and implement them within the single-image setting.
Our tuning-free comparison covers all available open-source methods, they are ELITE~\citep{elite}, BLIP-Diffusion~\citep{blipdiffusion}, and IP-Adapter~\citep{ip-adapter}.
As demonstrated in Tab.~\ref{tab:metrics}, our approach achieves the highest text-alignment score (C-T), indicating its effectiveness in eliminating subject-irrelevant information and significantly enhancing editability.
Although IP-Adapter~\citep{ip-adapter} and DreamBooth~\citep{dreambooth} achieve high image-alignment scores (C-I and D-I), they struggle with editability. This is because they tend to replicate large portions of the reference image, resulting in excessive consistency and reduced flexibility in making edits (Fig.~\ref{fig:more_result1} and~\ref{fig:more_result2}).
Excluding these two methods, our approach achieves the highest image-alignment scores (C-I and D-I), demonstrating its superior ID consistency without compromising text-alignment. 
We also achieve lower internal variance (IV) between the generated images created with the same subject under different conditions, further demonstrating that DisEnvisioner is unaffected by irrelevant factors, such as subject pose and its surroundings. 
In terms of efficiency, thanks to our lightweight and effective design, eliminating the test-time tuning, only 1.96s is required for each customized generation. 
We also conduct a user study, as detailed in the supplementary materials, which further demonstrates the superiority of DisEnvisioner.

\subsection{Qualitative Results}
\label{subsec:qualitive}
To obtain deeper insights of the DisEnvisioner, we visualize its synthesized images against the selected five prevailing methods
across three different images per subject.
Fig.~\ref{fig:more_result1} and~\ref{fig:more_result2} clearly demonstrates DisEnvisioner's superiority in producing high-quality, editable images with strong ID-consistency.
Notably, the consistency in animal postures alongside the minimal impact of irrelevant backgrounds across reference images under the same textual instruction, also showcase our robustness and resistance to subject-irrelevant elements, \emph{i.e.}, only the subject-essential attributes are extracted and preserved.
This excellence also applies to non-live objects, as seen in Fig.~\ref{fig:more_result2}, where exceptional customizations for subject the boot and sunglasses demonstrate DisEnvisioner's accurate focus on subject-essential features, leading to superior editability, ID-consistency, and the overall visual quality of the generated images.

\subsection{Ablation Study}
\label{subsec:ablation}

\begin{figure}[!t]
    \centering
    \includegraphics[width=\linewidth]{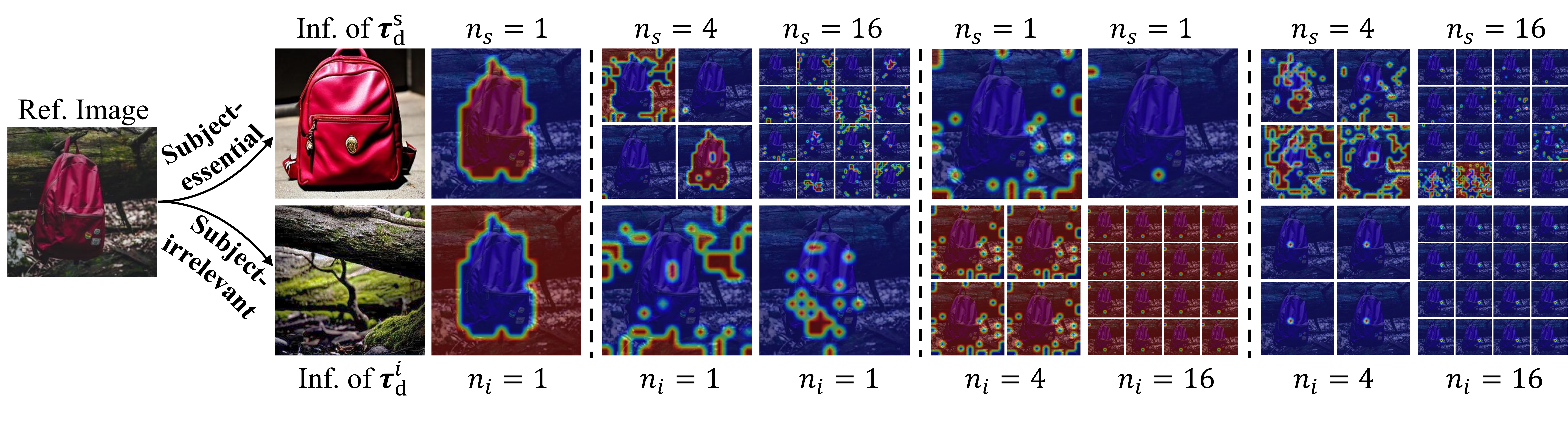}
    \caption{\textbf{Ablation on token's numbers in DisVisioner.}
    Each attention map is calculated by the dot product of the obtained token ($\tau^{{s}}_{\text{d}}, \tau^{{i}}_{\text{d}}$) and the CLIP local image features. 
    The results demonstrates that $n_s=1$ and $n_i=1$ outperforms other configurations, achieving precise attention map and faithful token inference. 
    }
    \label{fig:disentanglement}
\end{figure}


\begin{wrapfigure}{t}{0.3\textwidth}
\centering
\includegraphics[width = 1.0\linewidth]{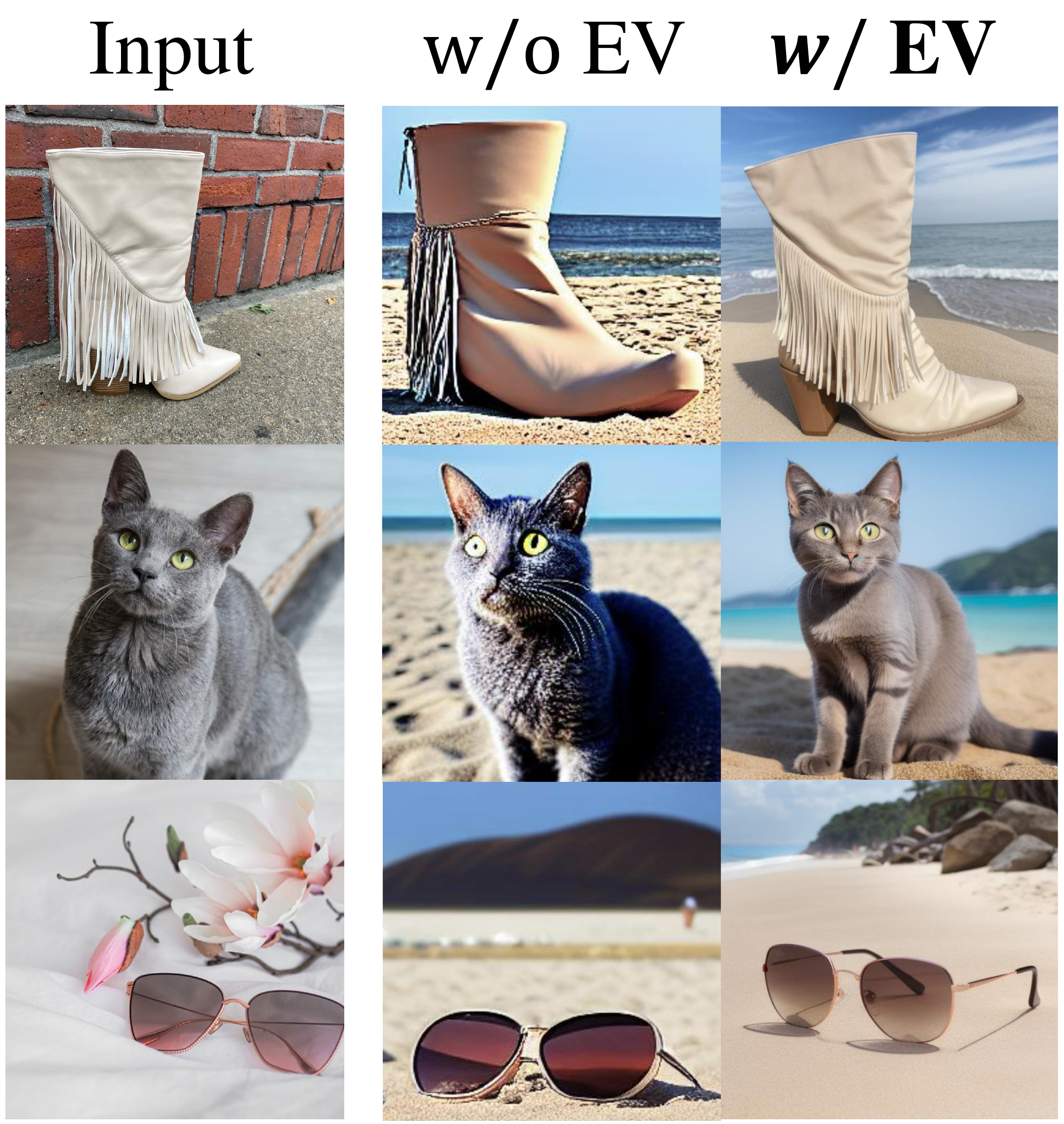}
\caption{\textbf{Ablation of EV.}}
\label{fig:abl_ev}
\end{wrapfigure}

\subsubsection{Influence of Tokens Numbers in DisVisioner.}
As described in Sec.~\ref{subsec:disvisioner}, the disentangled tokens are injected into the textual space of CLIP encoder, thus each token tends to represent a complete semantic meaning, similar to ``words''. Meanwhile, due to the mutual exclusivity between subject-essential and -irrelevant features in DisVisioner, an excessive number of tokens leads to increased competition, making disentanglement difficult and chaotic. For instance, in \cite{wu2022learning}, only four tokens are used to represent a complex outdoor scene, suggesting that fewer tokens may be more effective for simpler subjects. 
To achieve semantic completeness and avoid excessive feature competition, we conduct experiments on the number of subject-essential and -irrelevant tokens, \emph{i.e.}, $n_s$ and $n_i$. 
As visualized in Fig.~\ref{fig:disentanglement}, when $n_s\neq n_i$, the image is represented by the larger set without any disentanglement.
When $n_s = n_i>1$, random initialized subject tokens have more ``freedom'', capturing most image information and also leading to reconstruction. We also observe that all $n_i$ tokens are the same, that is because they use the same prior discussed in Sec.~\ref{subsec:disvisioner}. In summary, $n_i=1$ and $n_s=1$ performs the best. 

\subsubsection{Effect of EnVisioner.}
As shown in Fig.~\ref{fig:abl_ev}, we validate the effect of enriched subject representation in EnVisioner (abbreviated as EV). The prompt is ``\emph{a photo of S* on the beach}''. 
We can observe that the images enhanced by EnVisioner exhibit finer details and improved ID-consistency, with a significant boost in overall image quality,
emphasizing its crucial role in delivering high-quality customization.


\section{Conclusion}
\label{sec:conclusion}
In this paper, we propose \textbf{DisEnvisioner}, which is characterized by its emphasis on the interpretation of subject-essential attributes for high-quality image customization.
DisEnvisioner effectively identifies and enhances the subject-essential feature while filtering out other irrelevant information, enabling exceptional image customization without cumbersome tuning or relying on multiple reference images.
Through both the quantitative and qualitative evaluations, alongside the user study, we demonstrate DisEnvisioner' superior performance in customization quality and efficient inference time, offering a promising solution for practical applications.



\newpage
{\LARGE\sc Supplementary Materials of \\ \textbf{DisEnvisioner}:~\underline{Dis}entangled and~\underline{En}riched~\underline{Visual} Prompt for Customized Image Generation\par}

\appendix
\section{User Study}
We also conduct a user study using a round-robin format, where participants grade each method given the generation outputs across 3 images of the same subject in each round. There are 5 rounds in total. 
The metrics are text alignment (\textbf{TA}), subject identity alignment (\textbf{IA}), internal variance of those 3 customized images (\textbf{IV}), and image quality (\textbf{IQ}). A total of 69 users, and 345 rounds are recorded. As detailed in Table~\ref{tab:userstudy}, DisEnvisioner outperforms all baselines in all metrics. 

\begin{table}[h]
\footnotesize
\centering
\caption{\textbf{User study.} Participants rank the methods based on four criteria using a round-robin format, with the final scores are normalized before being reported. \textbf{mRank} is also reported.
}
\setlength{\tabcolsep}{11.pt}
\begin{tabular}{l|cccc|c}
\toprule
Method & TA$\uparrow$ & IA$\uparrow$ & IV$\uparrow$ & IQ $\uparrow$ & \textbf{mRank}$\downarrow$ \\
\midrule
DisenBooth~\citep{disenbooth}  &0.157 & 0.143& 0.156& 0.140&5.3\\
DreamBooth~\citep{dreambooth}        & 0.159&0.152 & 0.159& 0.158&3.5\\
\midrule
ELITE~\citep{elite}            & \underline{0.161}&0.147 & 0.154&0.147 &4.3\\
IP-Adapter~\citep{ip-adapter}      &0.153 &\underline{0.177} &0.152 &\underline{0.181}&4.0 \\
\underline{BLIP-Diffusion}~\citep{blipdiffusion}  &0.159 &0.162 & \underline{0.163}&0.165&\underline{2.8} \\
\midrule
\textbf{DisEnvisioner}& \textbf{0.211}& \textbf{0.219}& \textbf{0.216}&\textbf{0.209} &\textbf{1.0}    \\
\bottomrule
\end{tabular}
\label{tab:userstudy}
\end{table}

During each round of the user study, rather than \textbf{ranking} DisEnvisioner and five other existing methods from the best to the worst, users are expected to \textbf{assign grades} from 0 to 5 to each method according to specific metrics. 
In practice, all participants have the complete freedom to grade any method with any score based on their personal judgment.
After a total of 345 rounds of evaluation, the best-performing method often receives the highest scores, while scores for other methods are frequently identical due to similar customization quality. Additionally, it is uncommon for users to assign scores as low as 0 or 1.
Although the grading differences among methods are not particularly large, DisEnvisioner consistently outperforms others competitors across all evaluation criteria.

\section{Effect of $\lambda_s$ and $\lambda_i$}

\begin{figure}[t]
    \centering
    \includegraphics[width=0.9\linewidth]{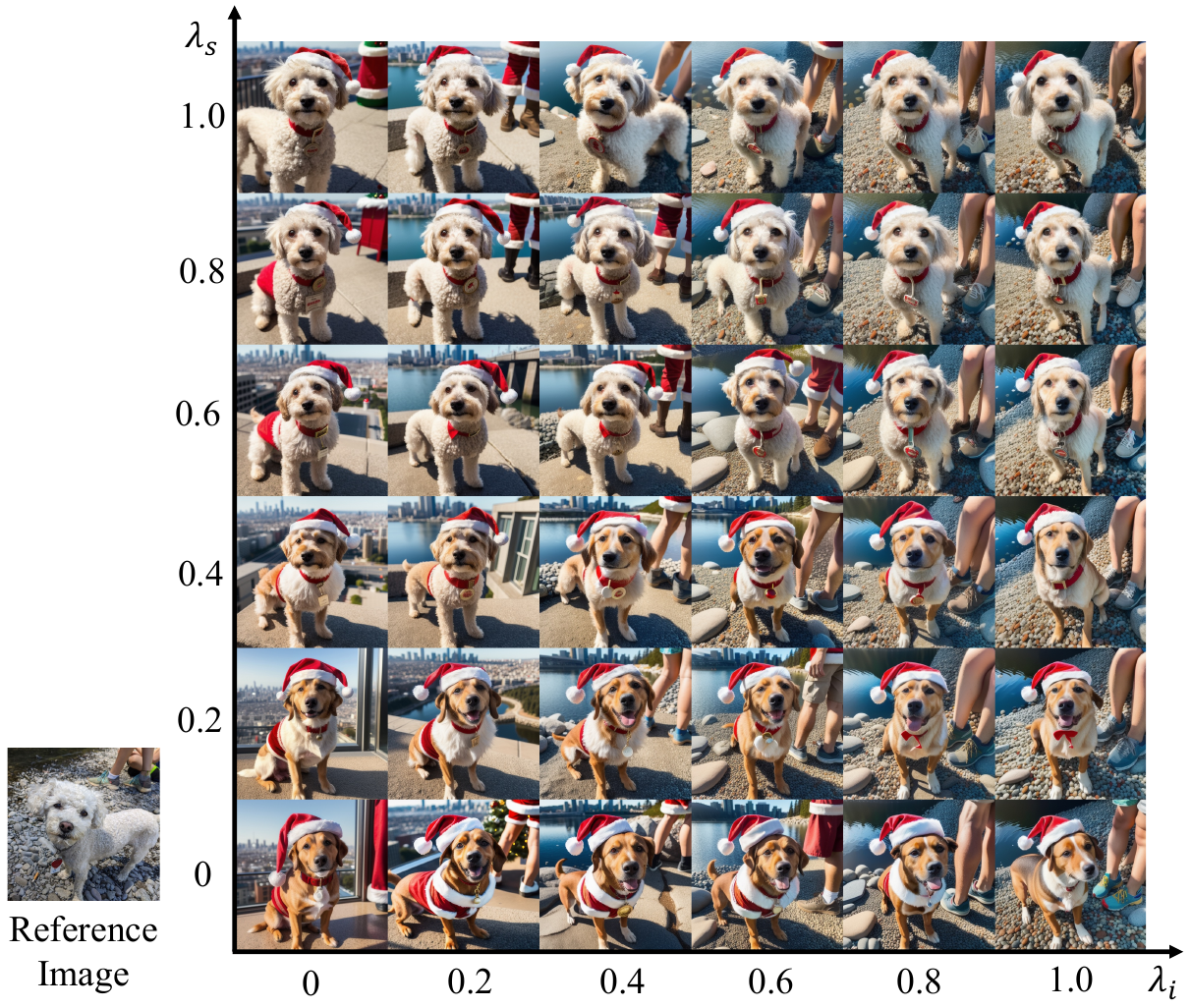}
    \caption{\textbf{Effect of varying $\lambda_s$ and $\lambda_i$} with class name provided. The prompt is ``a dog is wearing a Santa hat with a city in the background''.}
    \label{fig:lambda2}
\end{figure}

\begin{wrapfigure}{t}{0.33\textwidth}
\centering
\includegraphics[width = 1.0\linewidth]{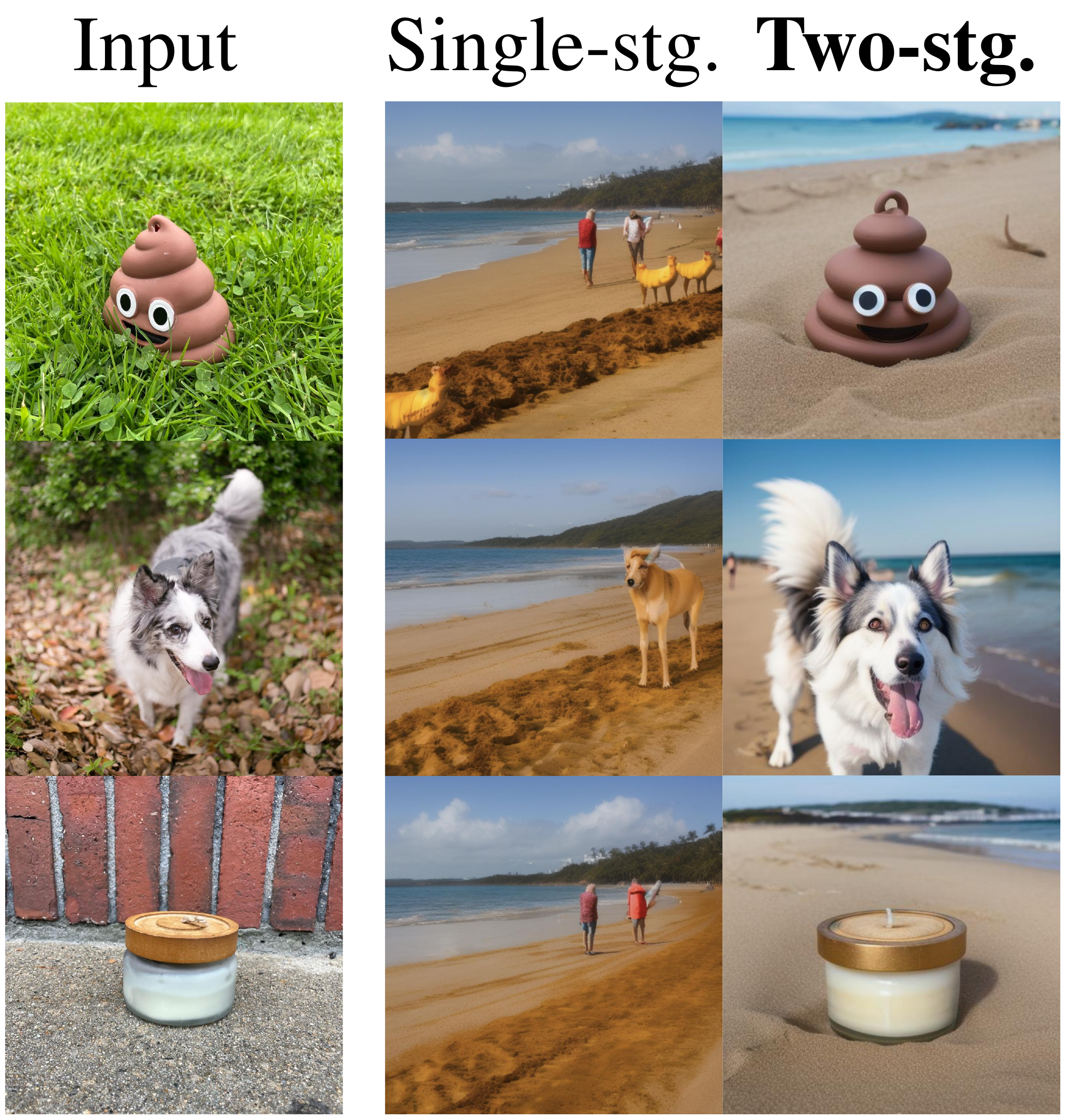}
\caption{\textbf{Comparison with single-stage and current two-stage training.} ``stg.'' represents the training stage. }
\label{fig:abl_srafe}
\end{wrapfigure}

As defined in Eq.~\ref{eq:envisioner} of the main paper, the weights $\lambda_s$ and $\lambda_i$ serve to modulate the integration of information that is essential and irrelevant to the subject in the given reference image. To thoroughly assess their effect of disentanglement, we adjust their values evenly sampled from 0 to 1.0 throughout the image customization process.

\begin{figure}[t]
    \centering
    \includegraphics[width=0.9\linewidth]{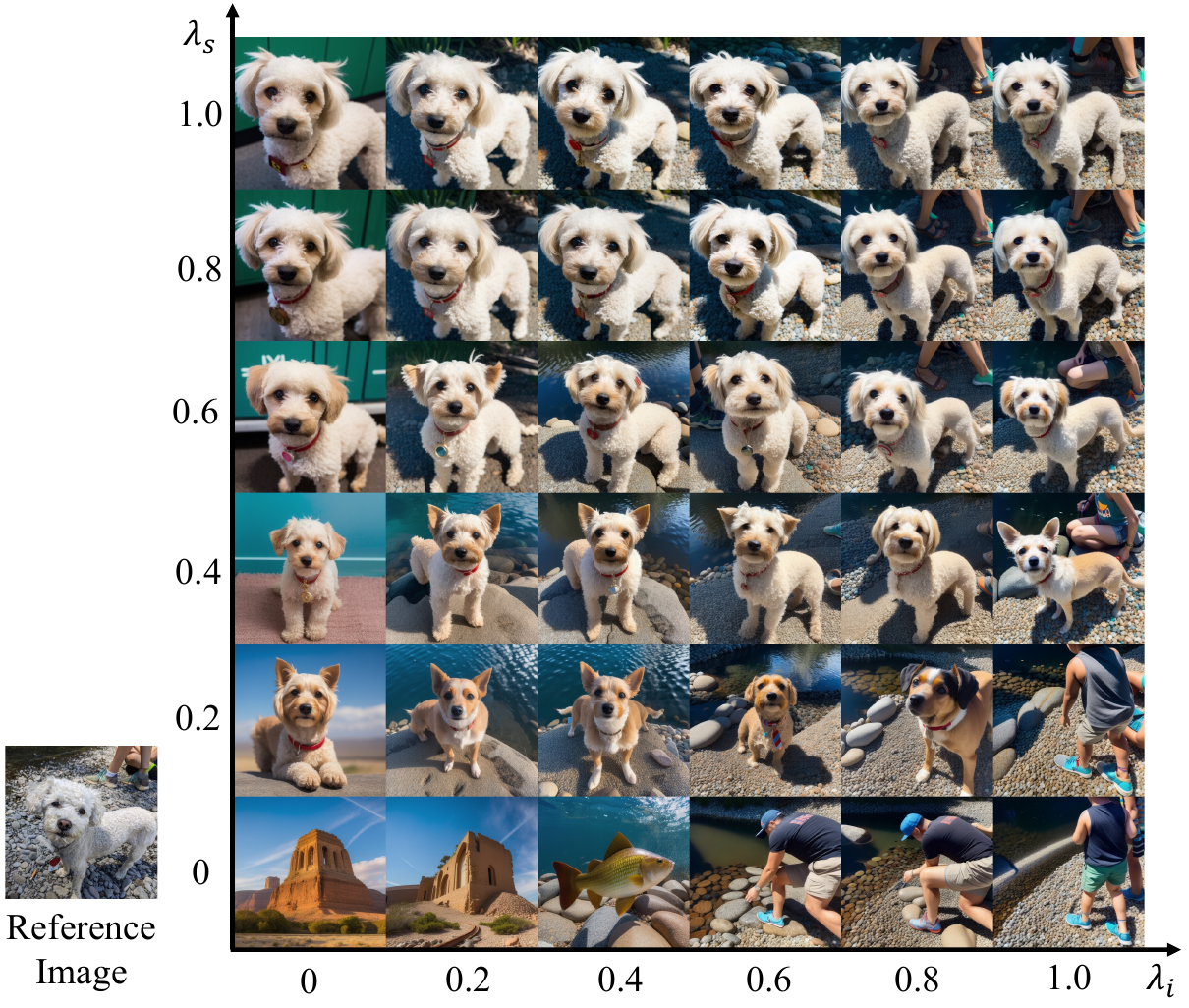}
    \caption{\textbf{Effect of varying $\lambda_s$ and $\lambda_i$} with \textit{empty} prompts without providing class names.}
    \label{fig:lambda1.1}
\end{figure}

\begin{figure}[t]
    \centering
    \includegraphics[width=0.9\linewidth]{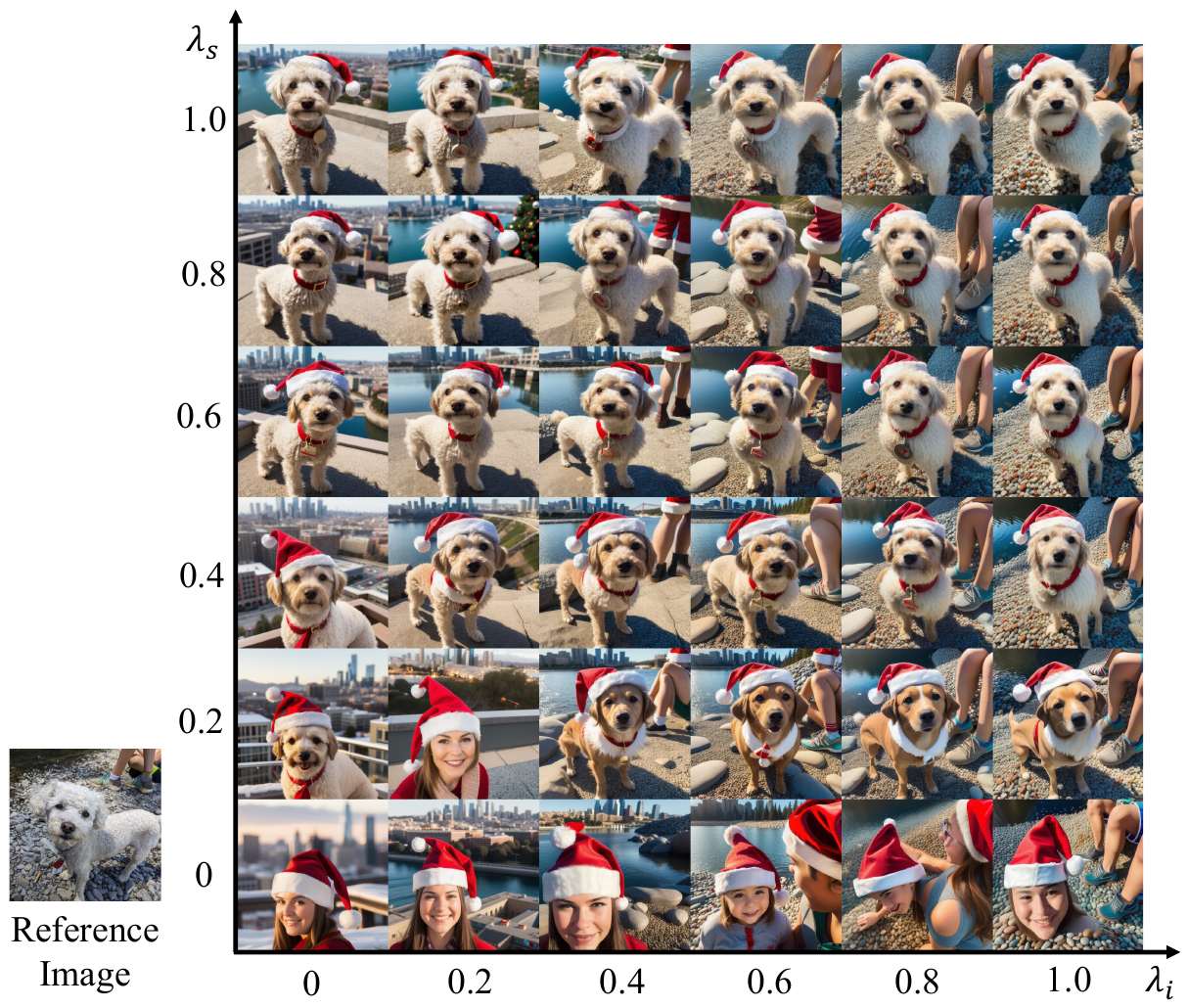} 
    \caption{\textbf{Effect of varying $\lambda_s$ and $\lambda_i$} with \textit{non-empty} prompts without providing class names. }
    \label{fig:lambda1.2}
\end{figure}

We generate images under varying settings of $\lambda_s$ and $\lambda_i$ employing both empty and non-empty (editing) prompts.
Fig.~\ref{fig:lambda1.1} and~\ref{fig:lambda1.2} demonstrate that as $\lambda_i$ decreases progressively (moving from the right to the left columns), the presence of subject-irrelevant disturbances in the images notably declines.
Also, enhancing $\lambda_s$ (moving from the bottom to the top rows) brings more pronounced consistency in subject identity between the reference and generated image.
When both $\lambda_s$ and $\lambda_i$ are reduced to their minimum value, \emph{i.e.}, $\lambda_s=0$ and $\lambda_i=0$, the generated images are solely influenced by the textual prompt, without incorporating any information from the reference image.
To further explore the role of additional information in the prompt, we also generate images with specific class names included in the prompts. 

As illustrated in Fig.~\ref{fig:lambda2} and Fig.~\ref{fig:lambda1.2}, particularly in terms of identity consistency, no matter the category-guidance (for instance, the class name ``dog'') is provided or not, it does not alter the customization quality.
This indicates that DisEnvisioner effectively deciphers and extracts subject-essential attributes from the reference image. By doing so, it can accurately identify relevant and redundant information, which is then eliminated.


It is also evident that when image generation focuses exclusively on solely subject-essential features ($\lambda_s=1.0$ and $\lambda_i=0$) or solely on purely subject-irrelevant features ($\lambda_s=0$ and $\lambda_i=1.0$), the reproduction of the subject and the irrelevant surrounding content is achieved independently, devoid of any interference from one another.
This phenomenon confirms the proficiency of DisEnvisioner in precisely segregating and enriching subject-essential features. It highlights the DisEnvisioner's exceptional customization performance without the need for test-time tuning, and relying solely on a single reference image.

\section{Can DisEnvisioner Be Trained in Single-stage?}

For single-stage training, we combine the training processes of DisVisioner and EnVisioner, aiming for simultaneous learning the disentanglement and enrichment of subject-essential features. However, in Fig.~\ref{fig:abl_srafe}, the single-stage model is difficult to be trained. It fails to capture any subject-essential information from the visual prompt and merely response to text instructions.
Nonetheless, the two-stage strategy of DisEnvisioner\textbf{---}separating disentanglement and enrichment\textbf{---}proves to be much more effective in high-quality customization.

\section{More Visualizations of Feature Disentanglement}
\begin{figure}[!t]
    \centering
    \includegraphics[width=0.8\textwidth]{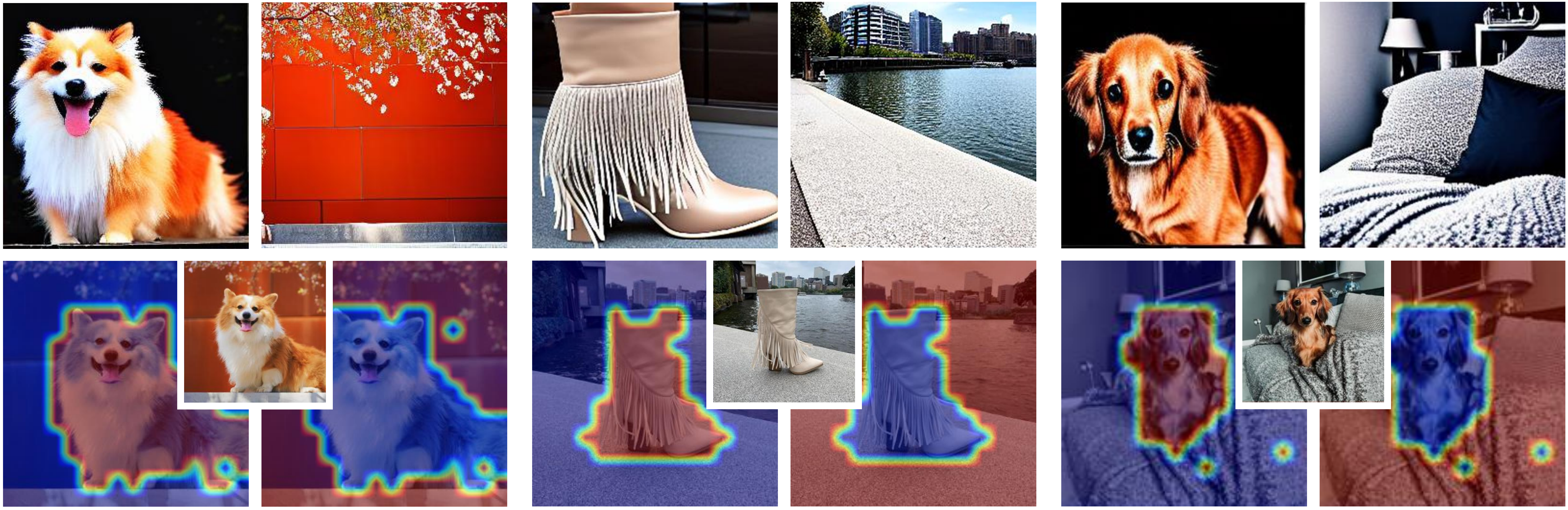}
    \caption{
    \textbf{More visualizations of feature disentanglement.} The results showcase DisVisioner’s ability to accurately discern subject-essential attributes across diverse scenarios. 
    }
    \label{fig:more_disen}
\end{figure}

As shown in Fig. \ref{fig:more_disen}, we provide more visualizations of feature disentanglement, across live and non-live subjects. With the $n_s=1$ and $n_i=1$, DisVisioner can clearly disentangle subject-essential and -irrelevant features, achieving more accurate image customization quality in diverse scenarios. 

\section{More Experimental Details}
\subsection{Training Data}
In the experiments, we utilize the \emph{training set} of OpenImages V6~\citep{kuznetsova2020open} as the training dataset. Based on this dataset, we construct \{prompt, image\} pairs for training. As depicted in Fig.~\ref{fig:training-data}, the training images are derived by cropping and resizing the raw images in accordance with the bounding box annotations. To ensure the quality of the training images, we further filter the cropped images. A cropped image is considered as unsatisfactory and therefore excluded if its area is greater than 80\% or less than 2\% of the original image's area. As a result, out of 14.61M annotated bounding boxes, we obtain 6.82M \{prompt, image\} pairs. The text prompts are selected randomly from a CLIP ImageNet template~\citep{radford2021learning} and integrated with labelled class names. The complete list of CLIP templates is provided below:
\begin{figure}[t]
    \centering
    \includegraphics[width=0.8\linewidth]{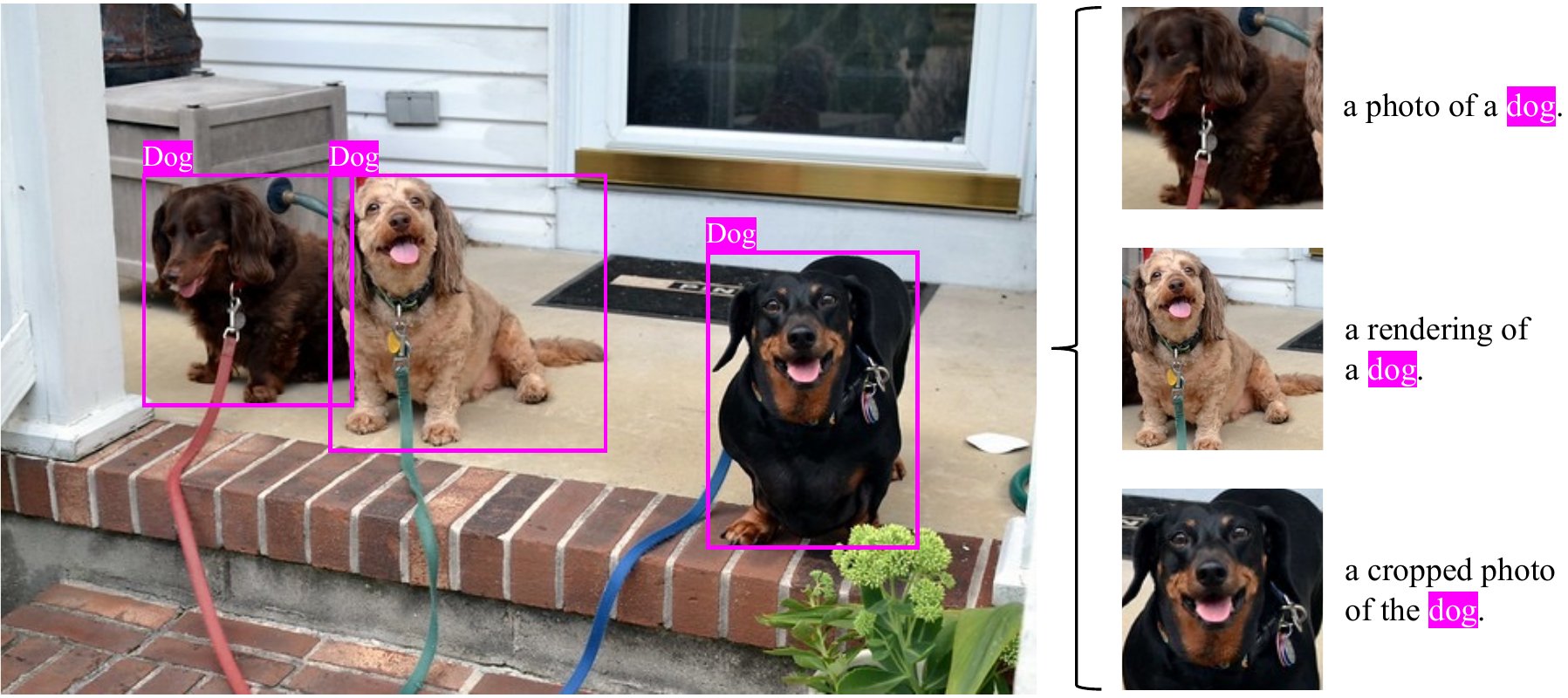}
    \caption{\textbf{Examples of training data.} The training images are derived by cropping and resizing the raw images in accordance with the bounding box annotations. }
    \label{fig:training-data}
\end{figure}

\begin{itemize}
    \item ``a photo of a $S^*$'',
    \item ``a rendering of a $S^*$'',
    \item ``a cropped photo of the $S^*$'',
    \item ``the photo of a $S^*$'',
    \item ``a photo of a clean $S^*$'',
    \item ``a photo of a dirty $S^*$'',
    \item ``a dark photo of the $S^*$'',
    \item ``a photo of my $S^*$'',
    \item ``a photo of the cool $S^*$'',
    \item ``a close-up photo of a $S^*$'',
    \item ``a bright photo of the $S^*$'',
    \item ``a cropped photo of a $S^*$'',
    \item ``a photo of the $S^*$'',
    \item ``a good photo of the $S^*$'',
    \item ``a photo of one $S^*$'',
    \item ``a close-up photo of the $S^*$'',
    \item ``a rendition of the $S^*$'',
    \item ``a photo of the clean $S^*$'',
    \item ``a rendition of a $S^*$'',
    \item ``a photo of a nice $S^*$'',
    \item ``a good photo of a $S^*$'',
    \item ``a photo of the nice $S^*$'',
    \item ``a photo of the small $S^*$'',
    \item ``a photo of the weird $S^*$'',
    \item ``a photo of the large $S^*$'',
    \item ``a photo of a cool $S^*$'',
    \item ``a photo of a small $S^*$''
\end{itemize}

\subsection{Testing Data}
\label{subsec:testing_data}
For evaluation, we adopt all images and editing prompts from DreamBooth~\citep{dreambooth} dataset. It contains a total of 158 images spanning 30 diverse categories, including dog, cat, robot, boot, etc. Fig.~\ref{fig:testing_data} shows a subset of these images. The the complete set of editing prompts for live subjects is detailed below:
\begin{figure}[t]
    \centering
    \includegraphics[width=0.9\linewidth]{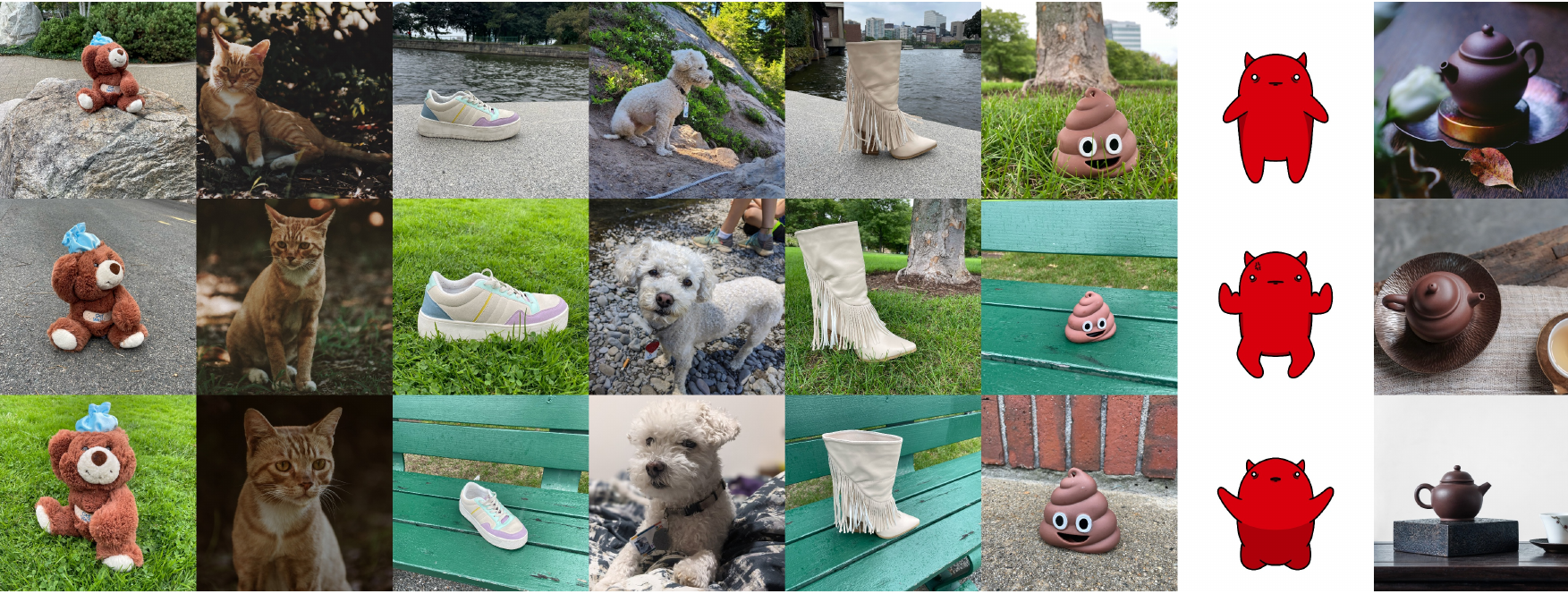}
    \caption{\textbf{Examples of testing data.} We selected all 30 subjects in the DreamBooth dataset.}
    \label{fig:testing_data}
\end{figure}

\begin{itemize}
\item ``a $S^*$ in the jungle''
\item ``a $S^*$ in the snow''
\item ``a $S^*$ on the beach''
\item ``a $S^*$ on a cobblestone street''
\item ``a $S^*$ on top of pink fabric''
\item ``a $S^*$ on top of a wooden floor''
\item ``a $S^*$ with a city in the background''
\item ``a $S^*$ with a mountain in the background''
\item ``a $S^*$ with a blue house in the background''
\item ``a $S^*$ on top of a purple rug in a forest''
\item ``a $S^*$ with a wheat field in the background''
\item ``a $S^*$ with a tree and autumn leaves in the background''
\item ``a $S^*$ with the Eiffel Tower in the background''
\item ``a $S^*$ floating on top of water''
\item ``a $S^*$ floating in an ocean of milk''
\item ``a $S^*$ on top of green grass with sunflowers around it''
\item ``a $S^*$ on top of a mirror''
\item ``a $S^*$ on top of the sidewalk in a crowded street''
\item ``a $S^*$ on top of a dirt road''
\item ``a $S^*$ on top of a white rug''
\item ``a red $S^*$''
\item ``a purple $S^*$''
\item ``a shiny $S^*$''
\item ``a wet $S^*$''
\item ``a cube shaped $S^*$''
\end{itemize}

And here is the full set of editing prompts for non-live subjects:

\begin{itemize}
\item ``a $S^*$ in the jungle''
\item ``a $S^*$ in the snow''
\item ``a $S^*$ on the beach''
\item ``a $S^*$ on a cobblestone street''
\item ``a $S^*$ on top of pink fabric''
\item ``a $S^*$ on top of a wooden floor''
\item ``a $S^*$ with a city in the background''
\item ``a $S^*$ with a mountain in the background''
\item ``a $S^*$ with a blue house in the background''
\item ``a $S^*$ on top of a purple rug in a forest''
\item ``a $S^*$ wearing a red hat''
\item ``a $S^*$ wearing a Santa hat''
\item ``a $S^*$ wearing a rainbow scarf''
\item ``a $S^*$ wearing a black top hat and a monocle''
\item ``a $S^*$ in a chef outfit''
\item ``a $S^*$ in a firefighter outfit''
\item ``a $S^*$ in a police outfit''
\item ``a $S^*$ wearing pink glasses''
\item ``a $S^*$ wearing a yellow shirt''
\item ``a $S^*$ in a purple wizard outfit''
\item ``a red $S^*$''
\item ``a purple $S^*$''
\item ``a shiny $S^*$''
\item ``a wet $S^*$''
\item ``a cube shaped $S*$''
\end{itemize}

\bibliography{iclr2024_conference}

\begin{thebibliography}{45}
\providecommand{\natexlab}[1]{#1}
\providecommand{\url}[1]{\texttt{#1}}
\expandafter\ifx\csname urlstyle\endcsname\relax
  \providecommand{\doi}[1]{doi: #1}\else
  \providecommand{\doi}{doi: \begingroup \urlstyle{rm}\Url}\fi

\bibitem[Arar et~al.(2024)Arar, Voynov, Hertz, Avrahami, Fruchter, Pritch, Cohen-Or, and Shamir]{arar2024palp}
Moab Arar, Andrey Voynov, Amir Hertz, Omri Avrahami, Shlomi Fruchter, Yael Pritch, Daniel Cohen-Or, and Ariel Shamir.
\newblock Palp: Prompt aligned personalization of text-to-image models.
\newblock \emph{arXiv preprint arXiv:2401.06105}, 2024.

\bibitem[Caron et~al.(2021)Caron, Touvron, Misra, J{\'e}gou, Mairal, Bojanowski, and Joulin]{dino}
Mathilde Caron, Hugo Touvron, Ishan Misra, Herv{\'e} J{\'e}gou, Julien Mairal, Piotr Bojanowski, and Armand Joulin.
\newblock Emerging properties in self-supervised vision transformers.
\newblock In \emph{Proceedings of the IEEE/CVF international conference on computer vision}, pp.\  9650--9660, 2021.

\bibitem[Chen et~al.(2023{\natexlab{a}})Chen, Zhang, Wang, Duan, Zhou, and Zhu]{disenbooth}
Hong Chen, Yipeng Zhang, Xin Wang, Xuguang Duan, Yuwei Zhou, and Wenwu Zhu.
\newblock Disenbooth: Disentangled parameter-efficient tuning for subject-driven text-to-image generation.
\newblock \emph{arXiv preprint arXiv:2305.03374}, 2023{\natexlab{a}}.

\bibitem[Chen et~al.(2023{\natexlab{b}})Chen, Yu, Ge, Yao, Xie, Wu, Wang, Kwok, Luo, Lu, et~al.]{pixart}
Junsong Chen, Jincheng Yu, Chongjian Ge, Lewei Yao, Enze Xie, Yue Wu, Zhongdao Wang, James Kwok, Ping Luo, Huchuan Lu, et~al.
\newblock Pixart-$\alpha$: Fast training of diffusion transformer for photorealistic text-to-image synthesis.
\newblock \emph{arXiv preprint arXiv:2310.00426}, 2023{\natexlab{b}}.

\bibitem[Chen et~al.(2023{\natexlab{c}})Chen, Zhao, Liu, Ding, Song, Wang, Wang, Yang, Liu, Du, et~al.]{photoverse}
Li~Chen, Mengyi Zhao, Yiheng Liu, Mingxu Ding, Yangyang Song, Shizun Wang, Xu~Wang, Hao Yang, Jing Liu, Kang Du, et~al.
\newblock Photoverse: Tuning-free image customization with text-to-image diffusion models.
\newblock \emph{arXiv preprint arXiv:2309.05793}, 2023{\natexlab{c}}.

\bibitem[Dhariwal \& Nichol(2021)Dhariwal and Nichol]{dhariwal2021diffusion}
Prafulla Dhariwal and Alexander Nichol.
\newblock Diffusion models beat gans on image synthesis.
\newblock \emph{Advances in neural information processing systems}, 34:\penalty0 8780--8794, 2021.

\bibitem[Dong et~al.(2022)Dong, Wei, and Lin]{dreamartist}
Ziyi Dong, Pengxu Wei, and Liang Lin.
\newblock Dreamartist: Towards controllable one-shot text-to-image generation via contrastive prompt-tuning.
\newblock \emph{arXiv preprint arXiv:2211.11337}, 2022.

\bibitem[Gal et~al.(2022)Gal, Alaluf, Atzmon, Patashnik, Bermano, Chechik, and Cohen-Or]{TI}
Rinon Gal, Yuval Alaluf, Yuval Atzmon, Or~Patashnik, Amit~H Bermano, Gal Chechik, and Daniel Cohen-Or.
\newblock An image is worth one word: Personalizing text-to-image generation using textual inversion.
\newblock \emph{arXiv preprint arXiv:2208.01618}, 2022.

\bibitem[Goodfellow et~al.(2014)Goodfellow, Pouget-Abadie, Mirza, Xu, Warde-Farley, Ozair, Courville, and Bengio]{goodfellow2014generative}
Ian Goodfellow, Jean Pouget-Abadie, Mehdi Mirza, Bing Xu, David Warde-Farley, Sherjil Ozair, Aaron Courville, and Yoshua Bengio.
\newblock Generative adversarial nets.
\newblock \emph{Advances in neural information processing systems}, 27, 2014.

\bibitem[Han et~al.(2023)Han, Li, Zhang, Milanfar, Metaxas, and Yang]{svdiff}
Ligong Han, Yinxiao Li, Han Zhang, Peyman Milanfar, Dimitris Metaxas, and Feng Yang.
\newblock Svdiff: Compact parameter space for diffusion fine-tuning.
\newblock \emph{arXiv preprint arXiv:2303.11305}, 2023.

\bibitem[He et~al.(2022)He, Zhou, Zhang, Peng, Shen, Sun, Chen, and Ji]{pixelfolder}
Jing He, Yiyi Zhou, Qi~Zhang, Jun Peng, Yunhang Shen, Xiaoshuai Sun, Chao Chen, and Rongrong Ji.
\newblock Pixelfolder: An efficient progressive pixel synthesis network for image generation.
\newblock \emph{arXiv preprint arXiv:2204.00833}, 2022.

\bibitem[He et~al.(2024)He, Li, Yin, Liang, Li, Zhou, Liu, Liu, and Chen]{he2024lotus}
Jing He, Haodong Li, Wei Yin, Yixun Liang, Leheng Li, Kaiqiang Zhou, Hongbo Liu, Bingbing Liu, and Ying-Cong Chen.
\newblock Lotus: Diffusion-based visual foundation model for high-quality dense prediction.
\newblock \emph{arXiv preprint arXiv:2409.18124}, 2024.

\bibitem[Ho et~al.(2020)Ho, Jain, and Abbeel]{ho2020denoising}
Jonathan Ho, Ajay Jain, and Pieter Abbeel.
\newblock Denoising diffusion probabilistic models.
\newblock \emph{Advances in neural information processing systems}, 33:\penalty0 6840--6851, 2020.

\bibitem[Hua et~al.(2023)Hua, Liu, Ding, Liu, Wu, and He]{hua2023dreamtuner}
Miao Hua, Jiawei Liu, Fei Ding, Wei Liu, Jie Wu, and Qian He.
\newblock Dreamtuner: Single image is enough for subject-driven generation.
\newblock \emph{arXiv preprint arXiv:2312.13691}, 2023.

\bibitem[Jia et~al.(2023)Jia, Zhao, Chan, Li, Zhang, Gong, Hou, Wang, and Su]{tamingencoder}
Xuhui Jia, Yang Zhao, Kelvin~CK Chan, Yandong Li, Han Zhang, Boqing Gong, Tingbo Hou, Huisheng Wang, and Yu-Chuan Su.
\newblock Taming encoder for zero fine-tuning image customization with text-to-image diffusion models.
\newblock \emph{arXiv preprint arXiv:2304.02642}, 2023.

\bibitem[Karras et~al.(2019)Karras, Laine, and Aila]{StyleGAN1}
Tero Karras, Samuli Laine, and Timo Aila.
\newblock A style-based generator architecture for generative adversarial networks.
\newblock In \emph{Proceedings of the IEEE/CVF conference on computer vision and pattern recognition}, pp.\  4401--4410, 2019.

\bibitem[Karras et~al.(2020)Karras, Laine, Aittala, Hellsten, Lehtinen, and Aila]{StyleGAN2}
Tero Karras, Samuli Laine, Miika Aittala, Janne Hellsten, Jaakko Lehtinen, and Timo Aila.
\newblock Analyzing and improving the image quality of stylegan.
\newblock In \emph{Proceedings of the IEEE/CVF conference on computer vision and pattern recognition}, pp.\  8110--8119, 2020.

\bibitem[Karras et~al.(2021)Karras, Aittala, Laine, H{\"a}rk{\"o}nen, Hellsten, Lehtinen, and Aila]{StyleGAN3}
Tero Karras, Miika Aittala, Samuli Laine, Erik H{\"a}rk{\"o}nen, Janne Hellsten, Jaakko Lehtinen, and Timo Aila.
\newblock Alias-free generative adversarial networks.
\newblock \emph{Advances in Neural Information Processing Systems}, 34:\penalty0 852--863, 2021.

\bibitem[Kumari et~al.(2023)Kumari, Zhang, Zhang, Shechtman, and Zhu]{customdiffusion}
Nupur Kumari, Bingliang Zhang, Richard Zhang, Eli Shechtman, and Jun-Yan Zhu.
\newblock Multi-concept customization of text-to-image diffusion.
\newblock In \emph{Proceedings of the IEEE/CVF Conference on Computer Vision and Pattern Recognition}, pp.\  1931--1941, 2023.

\bibitem[Kuznetsova et~al.(2020)Kuznetsova, Rom, Alldrin, Uijlings, Krasin, Pont-Tuset, Kamali, Popov, Malloci, Kolesnikov, et~al.]{kuznetsova2020open}
Alina Kuznetsova, Hassan Rom, Neil Alldrin, Jasper Uijlings, Ivan Krasin, Jordi Pont-Tuset, Shahab Kamali, Stefan Popov, Matteo Malloci, Alexander Kolesnikov, et~al.
\newblock The open images dataset v4: Unified image classification, object detection, and visual relationship detection at scale.
\newblock \emph{International Journal of Computer Vision}, 128\penalty0 (7):\penalty0 1956--1981, 2020.

\bibitem[Li et~al.(2024)Li, Li, and Hoi]{blipdiffusion}
Dongxu Li, Junnan Li, and Steven Hoi.
\newblock Blip-diffusion: Pre-trained subject representation for controllable text-to-image generation and editing.
\newblock \emph{Advances in Neural Information Processing Systems}, 36, 2024.

\bibitem[Li et~al.(2023{\natexlab{a}})Li, Li, Savarese, and Hoi]{li2023blip}
Junnan Li, Dongxu Li, Silvio Savarese, and Steven Hoi.
\newblock Blip-2: Bootstrapping language-image pre-training with frozen image encoders and large language models.
\newblock \emph{arXiv preprint arXiv:2301.12597}, 2023{\natexlab{a}}.

\bibitem[Li et~al.(2023{\natexlab{b}})Li, Wang, Duan, and Li]{li2023clip}
Yi~Li, Hualiang Wang, Yiqun Duan, and Xiaomeng Li.
\newblock Clip surgery for better explainability with enhancement in open-vocabulary tasks.
\newblock \emph{arXiv preprint arXiv:2304.05653}, 2023{\natexlab{b}}.

\bibitem[Li et~al.(2023{\natexlab{c}})Li, Cao, Wang, Qi, Cheng, and Shan]{photomaker}
Zhen Li, Mingdeng Cao, Xintao Wang, Zhongang Qi, Ming-Ming Cheng, and Ying Shan.
\newblock Photomaker: Customizing realistic human photos via stacked id embedding.
\newblock \emph{arXiv preprint arXiv:2312.04461}, 2023{\natexlab{c}}.

\bibitem[Loshchilov \& Hutter(2017)Loshchilov and Hutter]{loshchilov2017decoupled}
Ilya Loshchilov and Frank Hutter.
\newblock Decoupled weight decay regularization.
\newblock \emph{arXiv preprint arXiv:1711.05101}, 2017.

\bibitem[Ma et~al.(2023)Ma, Liang, Chen, and Lu]{subjectdiffusion}
Jian Ma, Junhao Liang, Chen Chen, and Haonan Lu.
\newblock Subject-diffusion: Open domain personalized text-to-image generation without test-time fine-tuning.
\newblock \emph{arXiv preprint arXiv:2307.11410}, 2023.

\bibitem[Nichol et~al.(2021)Nichol, Dhariwal, Ramesh, Shyam, Mishkin, McGrew, Sutskever, and Chen]{nichol2021glide}
Alex Nichol, Prafulla Dhariwal, Aditya Ramesh, Pranav Shyam, Pamela Mishkin, Bob McGrew, Ilya Sutskever, and Mark Chen.
\newblock Glide: Towards photorealistic image generation and editing with text-guided diffusion models.
\newblock \emph{arXiv preprint arXiv:2112.10741}, 2021.

\bibitem[Radford et~al.(2021)Radford, Kim, Hallacy, Ramesh, Goh, Agarwal, Sastry, Askell, Mishkin, Clark, et~al.]{radford2021learning}
Alec Radford, Jong~Wook Kim, Chris Hallacy, Aditya Ramesh, Gabriel Goh, Sandhini Agarwal, Girish Sastry, Amanda Askell, Pamela Mishkin, Jack Clark, et~al.
\newblock Learning transferable visual models from natural language supervision.
\newblock In \emph{International conference on machine learning}, pp.\  8748--8763. PMLR, 2021.

\bibitem[Ramesh et~al.(2021)Ramesh, Pavlov, Goh, Gray, Voss, Radford, Chen, and Sutskever]{dalle}
Aditya Ramesh, Mikhail Pavlov, Gabriel Goh, Scott Gray, Chelsea Voss, Alec Radford, Mark Chen, and Ilya Sutskever.
\newblock Zero-shot text-to-image generation.
\newblock In \emph{International Conference on Machine Learning}, pp.\  8821--8831. PMLR, 2021.

\bibitem[Ramesh et~al.(2022)Ramesh, Dhariwal, Nichol, Chu, and Chen]{unclip}
Aditya Ramesh, Prafulla Dhariwal, Alex Nichol, Casey Chu, and Mark Chen.
\newblock Hierarchical text-conditional image generation with clip latents.
\newblock \emph{arXiv preprint arXiv:2204.06125}, 1\penalty0 (2):\penalty0 3, 2022.

\bibitem[Rombach et~al.(2022)Rombach, Blattmann, Lorenz, Esser, and Ommer]{stablediffusion}
Robin Rombach, Andreas Blattmann, Dominik Lorenz, Patrick Esser, and Bj{\"o}rn Ommer.
\newblock High-resolution image synthesis with latent diffusion models.
\newblock In \emph{Proceedings of the IEEE/CVF conference on computer vision and pattern recognition}, pp.\  10684--10695, 2022.

\bibitem[Ronneberger et~al.(2015)Ronneberger, Fischer, and Brox]{ronneberger2015u}
Olaf Ronneberger, Philipp Fischer, and Thomas Brox.
\newblock U-net: Convolutional networks for biomedical image segmentation.
\newblock In \emph{Medical Image Computing and Computer-Assisted Intervention--MICCAI 2015: 18th International Conference, Munich, Germany, October 5-9, 2015, Proceedings, Part III 18}, pp.\  234--241. Springer, 2015.

\bibitem[Ruiz et~al.(2023)Ruiz, Li, Jampani, Pritch, Rubinstein, and Aberman]{dreambooth}
Nataniel Ruiz, Yuanzhen Li, Varun Jampani, Yael Pritch, Michael Rubinstein, and Kfir Aberman.
\newblock Dreambooth: Fine tuning text-to-image diffusion models for subject-driven generation.
\newblock In \emph{Proceedings of the IEEE/CVF Conference on Computer Vision and Pattern Recognition}, pp.\  22500--22510, 2023.

\bibitem[Saharia et~al.(2022)Saharia, Chan, Saxena, Li, Whang, Denton, Ghasemipour, Gontijo~Lopes, Karagol~Ayan, Salimans, et~al.]{imagen}
Chitwan Saharia, William Chan, Saurabh Saxena, Lala Li, Jay Whang, Emily~L Denton, Kamyar Ghasemipour, Raphael Gontijo~Lopes, Burcu Karagol~Ayan, Tim Salimans, et~al.
\newblock Photorealistic text-to-image diffusion models with deep language understanding.
\newblock \emph{Advances in Neural Information Processing Systems}, 35:\penalty0 36479--36494, 2022.

\bibitem[Shi et~al.(2023)Shi, Xiong, Lin, and Jung]{instantbooth}
Jing Shi, Wei Xiong, Zhe Lin, and Hyun~Joon Jung.
\newblock Instantbooth: Personalized text-to-image generation without test-time finetuning.
\newblock \emph{arXiv preprint arXiv:2304.03411}, 2023.

\bibitem[Song et~al.(2020)Song, Meng, and Ermon]{song2020denoising}
Jiaming Song, Chenlin Meng, and Stefano Ermon.
\newblock Denoising diffusion implicit models.
\newblock \emph{arXiv preprint arXiv:2010.02502}, 2020.

\bibitem[Voynov et~al.(2023)Voynov, Chu, Cohen-Or, and Aberman]{voynov2023p+}
Andrey Voynov, Qinghao Chu, Daniel Cohen-Or, and Kfir Aberman.
\newblock $ p+ $: Extended textual conditioning in text-to-image generation.
\newblock \emph{arXiv preprint arXiv:2303.09522}, 2023.

\bibitem[Wang et~al.(2024)Wang, Bai, Wang, Qin, and Chen]{instantid}
Qixun Wang, Xu~Bai, Haofan Wang, Zekui Qin, and Anthony Chen.
\newblock Instantid: Zero-shot identity-preserving generation in seconds.
\newblock \emph{arXiv preprint arXiv:2401.07519}, 2024.

\bibitem[Wei et~al.(2023)Wei, Zhang, Ji, Bai, Zhang, and Zuo]{elite}
Yuxiang Wei, Yabo Zhang, Zhilong Ji, Jinfeng Bai, Lei Zhang, and Wangmeng Zuo.
\newblock Elite: Encoding visual concepts into textual embeddings for customized text-to-image generation.
\newblock \emph{arXiv preprint arXiv:2302.13848}, 2023.

\bibitem[Wu et~al.(2022)Wu, Wang, Zhou, Hu, and Li]{wu2022learning}
Hui Wu, Min Wang, Wengang Zhou, Yang Hu, and Houqiang Li.
\newblock Learning token-based representation for image retrieval.
\newblock In \emph{Proceedings of the AAAI Conference on Artificial Intelligence}, volume~36, pp.\  2703--2711, 2022.

\bibitem[Xu et~al.(2018)Xu, Zhang, Huang, Zhang, Gan, Huang, and He]{xu2018attngan}
Tao Xu, Pengchuan Zhang, Qiuyuan Huang, Han Zhang, Zhe Gan, Xiaolei Huang, and Xiaodong He.
\newblock Attngan: Fine-grained text to image generation with attentional generative adversarial networks.
\newblock In \emph{Proceedings of the IEEE conference on computer vision and pattern recognition}, pp.\  1316--1324, 2018.

\bibitem[Ye et~al.(2023)Ye, Zhang, Liu, Han, and Yang]{ip-adapter}
Hu~Ye, Jun Zhang, Sibo Liu, Xiao Han, and Wei Yang.
\newblock Ip-adapter: Text compatible image prompt adapter for text-to-image diffusion models.
\newblock 2023.

\bibitem[Zhang et~al.(2017)Zhang, Xu, Li, Zhang, Wang, Huang, and Metaxas]{zhang2017stackgan}
Han Zhang, Tao Xu, Hongsheng Li, Shaoting Zhang, Xiaogang Wang, Xiaolei Huang, and Dimitris~N Metaxas.
\newblock Stackgan: Text to photo-realistic image synthesis with stacked generative adversarial networks.
\newblock In \emph{Proceedings of the IEEE international conference on computer vision}, pp.\  5907--5915, 2017.

\bibitem[Zhang et~al.(2018)Zhang, Xu, Li, Zhang, Wang, Huang, and Metaxas]{zhang2018stackgan++}
Han Zhang, Tao Xu, Hongsheng Li, Shaoting Zhang, Xiaogang Wang, Xiaolei Huang, and Dimitris~N Metaxas.
\newblock Stackgan++: Realistic image synthesis with stacked generative adversarial networks.
\newblock \emph{IEEE transactions on pattern analysis and machine intelligence}, 41\penalty0 (8):\penalty0 1947--1962, 2018.

\bibitem[Zhang et~al.(2021)Zhang, Koh, Baldridge, Lee, and Yang]{zhang2021cross}
Han Zhang, Jing~Yu Koh, Jason Baldridge, Honglak Lee, and Yinfei Yang.
\newblock Cross-modal contrastive learning for text-to-image generation.
\newblock In \emph{Proceedings of the IEEE/CVF conference on computer vision and pattern recognition}, pp.\  833--842, 2021.

\end{thebibliography}
\bibliographystyle{iclr2024_conference}

\end{document}